\documentclass[Afour,sageh,times]{sagej}
\usepackage[acronym, toc, nonumberlist, shortcuts]{glossaries}
\usepackage{moreverb,url}

\usepackage[colorlinks,bookmarksopen,bookmarksnumbered,citecolor=red,urlcolor=red]{hyperref}

\usepackage{xspace}
\usepackage{dirtytalk}
\usepackage{siunitx}
\usepackage[nameinlink,noabbrev,capitalise]{cleveref}
\usepackage{graphicx}
\usepackage{float}
\usepackage{rotating}
\usepackage{caption}
\usepackage{subcaption}
\usepackage{psfrag}
\usepackage{amsmath,amssymb,amsfonts,bbm,nicefrac,mathtools,lipsum}
\usepackage{empheq}
\usepackage{multirow}
\usepackage{latexsym}
\usepackage{balance}
\usepackage{colortbl}
\usepackage[table]{xcolor}
\usepackage[super]{nth}
\usepackage{fancyhdr}
\usepackage{epstopdf}
\usepackage{float}
\usepackage{booktabs,tabularx}
\usepackage{makecell}
\usepackage{academicons}
\usepackage{cancel}
\usepackage{xparse}

\newtheorem{remark}{Remark}

\newcommand{\defeq}{\coloneqq}
\newcommand{\eg}{e.g.,\xspace}
\newcommand{\ie}{i.e.,\xspace}

\newacronym{em}{EM}{expectation maximization}
\newacronym{ppo}{PPO}{proximal policy optimization}
\newacronym{drl}{DRL}{deep reinforcement learning}
\newacronym{il}{IL}{imitation learning}
\newacronym{ece}{ECE}{entropic cost equilibrium}
\newacronym{ioc}{IOC}{inverse optimal control}
\newacronym{lqr}{LQR}{linear-quadratic regulator}
\newacronym{lqg}{LQG}{linear-quadratic-Gaussian}
\newacronym{ilqg}{ilqg}{iterative linear-quadratic Gaussian}
\newacronym{ilqr}{iLQR}{iterative linear-quadratic regulator}
\newacronym{kkt}{KKT}{Karush–Kuhn–Tucker}
\newacronym{irl}{IRL}{inverse reinforcement learning}
\newacronym{mle}{MLE}{maximum likelihood estimation}
\newacronym{map}{MAP}{maximum a posteriori}
\newacronym{sem}{SEM}{standard error of the mean}
\newacronym{mpgp}{MPGP}{model-predictive game-play}
\newacronym[longplural=open-loop Nash equilibria,\glsshortpluralkey={OLNE}]{olne}{OLNE}{open-loop Nash equilibrium}
\newacronym{licq}{LICQ}{linear independence constraint qualification}
\newacronym{mpc}{MPC}{model-predictive control}
\newacronym[longplural=mathematical programs with equilibrium constraints,plural={MPEC}]{mpec}{MPEC}{mathematical program with equilibrium constraints}
\newacronym[longplural={partially observable Markov decision processes}]{pomdp}{POMDP}{partially observable Markov decision process}
\newacronym{posg}{POSG}{partially observable stochastic game}
\newacronym{nep}{NEP}{Nash equilibrium problem}
\newacronym{gnep}{GNEP}{generalized Nash equilibrium problem}
\newacronym[longplural={generalized Nash equilibria},\glsshortpluralkey={GNE}]{gne}{GNE}{generalized Nash equilibrium}
\newacronym{svo}{SVO}{social value orientation}
\newacronym{ukf}{UKF}{unscented Kalman filter}
\newacronym{ibr}{IBR}{iterated best response}
\newacronym{awgn}{AWGN}{additive white Gaussian noise}
\newacronym{iqr}{IQR}{interquartile range}
\newacronym{mcp}{MCP}{mixed complementarity problem}
\newacronym{ift}{IFT}{implicit function theorem}
\newacronym{nn}{NN}{neural network}
\newacronym{rrt}{RRT}{rapidly exploring random trees}
\newacronym{idm}{IDM}{intelligent driver model}
\newacronym{lstm}{LSTM}{long short-term
memory}
\newacronym{lq}{LQ}{linear-quadratic}
\newacronym{gail}{GAIL}{generative adversarial imitation learning}
\newacronym{dagger}{DAgger}{dataset aggregation}
\newacronym{vae}{VAE}{variational autoencoder}
\newacronym{vi}{VI}{variational inference}
\newacronym{cgan}{(C)GAN}{(conditional) generative adversarial network}
\newacronym{cvae}{(C)VAE}{(conditional) variational autoencoder}
\newacronym{rnn}{RNN}{recurrent neural network}
\newacronym{fov}{FOV}{field-of-view}
\newacronym{ros}{ROS}{Robot Operating System}
\newacronym{kl}{KL}{Kullback–Leibler}
\newacronym{elbo}{ELBO}{evidence lower bound}
\newacronym{sgd}{SGD}{stochastic gradient descent}
\newacronym{birl}{BIRL}{Bayesian inverse reinforcement learning}
\newacronym{sqp}{SQP}{sequential quadratic programming}
\newacronym{ood}{OOD}{out-of-distribution}
\newacronym{stl}{STL}{signal temporal logic}
\newacronym{bpine}{B-PinE}{planning in expectation}
\newcommand{\observation}{y}
\newcommand{\observationTrajectorySpace}{{\reals^{\trajectoryObsDim}}}
\newcommand{\observationImageSpace}{{\reals^{\imageObsDim}}}
\newcommand{\observationSpace}{{\observationTrajectorySpace\times\observationImageSpace}}
\newcommand{\observedTrajectory}{{y_\text{traj}}}
\newcommand{\observedImage}{{y_\mathrm{img}}}
\newcommand{\dkl}[2]{D_\mathrm{KL}\left(#1 \parallel #2\right)}

\newcommand{\gameParams}{\theta}
\newcommand{\encoderParams}{\psi}
\newcommand{\decoderParams}{\phi}

\newcommand{\gameDecoder}[1][\decoderParams]{d^{\game}_{#1}}
\newcommand{\imageDecoder}[1][\decoderParams]{d^{\mathrm{img}}_{#1}}
\newcommand{\toTrajectoryObservationMean}{h_\text{traj}}

\newcommand{\latentSpace}{\reals^{d_z}}
\newcommand{\encoder}[1][\encoderParams]{e_{#1}}
\newcommand{\decoder}[1][\decoderParams]{d_{#1}}
\DeclareRobustCommand{\expectedValue}[2]{\mathbb{E}_{#1}\left[#2\right]}

\newcommand{\dataset}{\mathcal{D}}
\newcommand{\latent}{z}
\newcommand{\traj}{\tau}
\newcommand{\solveGame}{\mathcal{T}_{\game}}

\newcommand{\normaldist}{\mathcal{N}}
\newcommand{\transpose}{\top}
\newacronym{iid}{iid}{independent and identically distributed}
\newacronym{ji}{JI}{Jensen's inequality}
\newacronym{pdf}{PDF}{probability density function}

\newacronym{hj}{HJ}{Hamilton–Jacobi}

\newcommand{\given}{\mid}

\newcommand{\costparams}{\theta}

\newcommand{\game}{\Gamma}

\newcommand{\reals}{\mathbb{R}}

\newcommand{\revised}[1]{{\leavevmode\color{black}#1}}

\newcommand{\para}[1]{\smallskip\noindent\textbf{#1.}}

\newcommand{\lagrangian}{\mathcal{L}}
\newcommand{\dualMultiplier}{\lambda}
\newcommand{\activeConditions}{F_\Gamma}
\newcommand{\activeIndices}{\mathcal{I}}
\newcommand{\extractRows}[2]{{\left[#2\right]}_{#1}}
\newcommand{\primalsAndDuals}{v^*}

\newcommand{\trajectoryObsDim}{d_\observedTrajectory}
\newcommand{\imageObsDim}{d_\observedImage}

\defcitealias{MCPTrajectoryGameSolver}{JGTP Team, 2023}

\newcommand\BibTeX{{\rmfamily B\kern-.05em \textsc{i\kern-.025em b}\kern-.08em
T\kern-.1667em\lower.7ex\hbox{E}\kern-.125emX}}

\def\volumeyear{2016}

\begin{document}

\runninghead{Jain et al.}

\title{Bayesian Inverse Games with High-Dimensional Multi-Modal Observations}

\author{
Yash Jain\affilnum{1*},
Xinjie Liu\affilnum{1*},
Lasse Peters\affilnum{2*},
David Fridovich-Keil\affilnum{1}, and 
Ufuk Topcu\affilnum{1}
}

\affiliation{
\affilnum{*}Equal contribution\\
\affilnum{1}The University of Texas at Austin, Austin, TX 78712, USA\\
\affilnum{2}Delft University of Technology, Delft, 2600 AA, Netherlands
}

\corrauth{
Xinjie Liu,
The University of Texas at Austin, Austin, TX 78712, USA.
}

\email{xinjie-liu@utexas.edu}

\begin{abstract}

Many multi-agent interaction scenarios can be naturally modeled as noncooperative games, where each agent’s decisions depend on others’ future actions. 
However, deploying game-theoretic planners for autonomous decision-making requires a specification of \emph{all} agents' objectives. 
To circumvent this practical difficulty, recent work develops maximum likelihood techniques for solving \emph{inverse} games that can identify unknown agent objectives from interaction data. 
Unfortunately, these methods only infer point estimates and do not quantify estimator uncertainty; correspondingly, downstream planning decisions can overconfidently commit to unsafe actions.
We present an approximate Bayesian inference approach for solving the inverse game problem, which can incorporate observation data from multiple modalities and be used to generate samples from the Bayesian posterior over the hidden agent objectives given limited sensor observations in real time.
Concretely, the proposed Bayesian inverse game framework  trains a structured \acl{vae} with an embedded differentiable Nash game solver on interaction datasets and does not require labels of agents' true objectives. 
Extensive experiments show that our framework successfully learns prior and posterior distributions, improves inference quality over \acl{mle}-based inverse game approaches, and enables safer downstream decision-making without sacrificing efficiency. When trajectory information is uninformative or unavailable, multimodal inference further reduces uncertainty by exploiting additional observation modalities.

\end{abstract}


\maketitle

\section{Introduction}\label{sec:intro}

Noncooperative game theory~\citep{Basar1998games} provides a principled framework for modeling multi-agent interactive decision-making and has become a powerful tool for interaction-aware motion planning in autonomous robots.
Modern numerical methods~\citep{fridovich2020efficient,zhu2023sequential,cleac2022algames,li2023scenario} have significantly accelerated the solution of game-theoretic planning problems, enabling real-time application and increasing their practical appeal~\citep{spica2020real,wang2021tro}.

\begin{figure}[t]
  \centering
  \includegraphics[width=0.9\linewidth]{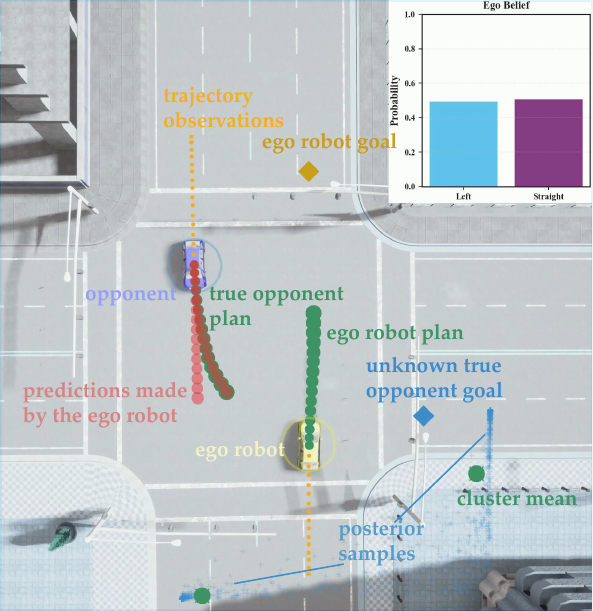}
  \caption{
A robot interacts with an opponent driver whose goal position is unknown. We embed a differentiable game solver within a structured variational autoencoder to infer the \emph{distribution} of opponent intent from observed trajectories, jointly with image observations of the opponent. 
}
  \label{fig:motivation}
\end{figure}

In game-theoretic planning, an ego agent (e.g., an autonomous robot) models its own and others’ decision-making processes as coupled optimization problems with potentially conflicting objectives. 
When each agent's decision is \emph{unilaterally} optimal, the joint decision is a Nash equilibrium. 
For example, \cref{fig:motivation} illustrates this type of noncooperative interaction in an intersection scenario between a red ego robot and a green opponent (e.g., a human driver), where they both seek to reach their goals efficiently while avoiding collisions.

In real-world deployments, however, such methods are typically used in a \emph{decentralized} fashion, where each agent only knows its own objective; thus, to compute interaction-aware Nash equilibrium motion plans, the ego robot must reason about \emph{unknown game parameters}, such as the objectives of other agents. In the example of \cref{fig:motivation}, 
to interact safely while still efficiently reaching its goal, the ego must, within split seconds, infer whether the opponent will go straight or turn left from their observed opponent behavior and, when available, visual cues such as turn signals.

The problem of inferring unknown game parameters from observations is known as an \emph{inverse game}~\citep{molloy2022inverse}. Standard approaches typically pose a \ac{mle} problem~\citep{peters_online_2023,liu2023learning,11217215,mehr2023maxent,hu2025think,sun2025inverse} with the first-order Nash equilibrium conditions as constraints. However, \ac{mle} yields inherently overconfident point estimates: it lacks uncertainty quantification and often ignores important prior knowledge, which can lead to unsafe decisions. For example, when an opponent has just entered the intersection and has not yet committed to a turning maneuver, an \ac{mle} approach may confidently conclude that the opponent will go straight, whereas an experienced human driver would remain cautious, understanding the potential for an aggressive left turn~\citep{liu2024auto}.

\emph{Bayesian inverse games} instead cast parameter inference as a Bayesian inference problem, which can be approximately solved by online filtering methods such as particle filtering~\citep{peters2020accommodating} or unscented Kalman filtering~\citep{le2021lucidgames}. More recently, \citet{liu2024auto} proposed a structured \ac{vae} framework that approximately solves the Bayesian inverse game problem via amortized, offline training on datasets of interaction trajectories without parameter labels. The resulting generative belief model infers posterior distributions over unknown opponent objectives in noncooperative games from observed trajectories, runs in real time, and captures multimodal posteriors, providing uncertainty quantification which improves downstream motion planning safety.

However, prior inverse game methods operate almost exclusively on low-dimensional observations, \ie state trajectories. In real-world interactions, other observation modalities—such as visual cues (turn signals, vehicle type, etc.)—can carry crucial information about an opponent’s intent, especially when a new agent has just entered the scene and little or no trajectory history is available. In highly dynamic, interactive settings, the ability to exploit such contextual information beyond pure behavior observations can be safety-critical, motivating inverse game approaches that operate on \emph{high-dimensional, multimodal observations} from multiple sensors.

To this end, this article substantially extends the conference paper by \citet{liu2024auto} (in press; to appear in \emph{Algorithmic Foundations of Robotics XVI}, Springer Nature) and contributes a tractable Bayesian inverse game framework that embeds a differentiable Nash game solver into a \ac{vae}. The proposed framework:
\begin{itemize}
    \item \textbf{(Tractability and interpretability)} leverages analytical gradients from a differentiable Nash game solver to learn a structured generative belief model, making multimodal posterior inference tractable while preserving interpretability for downstream motion planners;
    \item \textbf{(Multi-modal observation fusion)} fuses visual cues from raw images with partial-state observations of agent interactions to perform amortized Bayesian inference, reducing uncertainty when trajectory data are unavailable or uninformative;
    \item \textbf{(Downstream decision-making)} enables safer and more efficient uncertainty-aware decision-making than \ac{mle}-based inverse game approaches, while running in real time.
\end{itemize}

Beyond the conference version, the present work:
\begin{itemize}
    \item generalizes the original Bayesian inverse game framework and architecture to handle multiple observation modalities, including high-dimensional raw images;
    \item introduces new experimental studies in CARLA simulations~\citep{Dosovitskiy17} with photorealistic visual observations and low-level tracking controllers;
    \item systematically evaluates the impact of multimodal observations on Bayesian inference quality and downstream motion planning performance.
\end{itemize}

Throughout the manuscript, we use “multi-modal” to denote either (i) multi-peaked prior/posterior distributions or (ii) multiple observation modalities (e.g., trajectories and images); the intended meaning is clear from context.

\section{Related Work}\label{sec:related-works}

This section provides an overview of the literature on dynamic game theory, focusing on both forward games (\cref{related-works-forward-games}) and inverse games (\cref{related-works-inverse-games}).\footnote{\cref{related-works-forward-games,related-works-inverse-games} are largely reproduced from our earlier conference publication~\citep{liu2024auto}.}

\subsection{Dynamic Games}
\label{related-works-forward-games}
This work focuses on noncooperative dynamic games where agents can have arbitrary, potentially conflicting objectives, do not collude, and make decisions sequentially over time~\citep{Basar1998games}.
Since we assume that agents take actions simultaneously without leader-follower hierarchy and we consider coupling between agents' decisions through both objectives and constraints, our focus is on \acp{gnep}, for which a number of efficient computational methods have arisen in recent years.

Due to the computational challenges involved in solving such problems under feedback information structure~\citep{laine2021computation}, most works aim to find \acl{olne}~\citep{Basar1998games} instead, where players choose their entire action sequence---an open-loop strategy---at once. %
A substantial body of work employs the iterated best response algorithm to find a open-loop Nash equilibrium by iteratively solving single-agent optimization problems~%
\citep{williams2018best,spica2020real,wang2019game,spica2020real,wang2021tro,schwarting2019social}.
More recently, methods based on sequential quadratic approximations have been proposed~\citep{cleac2022algames,zhu2023sequential}, aiming to speed up convergence by updating all players' open-loop strategies simultaneously at each iteration.
Finally, since the first-order necessary conditions of open-loop \acp{gnep} take the form of a \ac{mcp}~\citep{mcp_ref}, several works~\citep{liu2023learning,peters2024ral} solve \acp{gne} using established \ac{mcp} solvers~\citep{billups1997comparison,dirkse1995path}.
This work builds on the latter approach.

\subsection{Inverse Dynamic Games}\label{related-works-inverse-games}
Inverse games study the problem of inferring unknown game parameters, e.g. of objective functions, from observations of agents' behavior~\citep{waugh2013computational}.
In recent years, several approaches have extended single-agent \ac{ioc} and \ac{irl} techniques to multi-agent interactive settings.
For instance, the approaches of~\citep{ROTHFU201714909, awasthi2020inverse} minimize the residual of agents' first-order necessary conditions, given \emph{full} state-control observations, in order to infer unknown objective parameters.
This approach is further extended to maximum-entropy settings in~\citep{inga2019inverse}.

Recent work~\citep{peters2023ijrr} proposes to maximize observation likelihood while enforcing the \ac{kkt} conditions of \acp{olne} as constraints.
This approach only requires partial-state observations and can cope with noise-corrupted data.
\citet{li2023cost} and \citet{liu2023learning} propose an extension of the \ac{mle} approach~\citep{peters2023ijrr} to inverse feedback and open-loop games with inequality constraints, exploiting the directional differentiability of generalized Nash equilibria with respect to problem parameters.
\revised{To amortize the computation of the \ac{mle},  \citet{geiger2021learning} and \citet{liu2023learning} demonstrate integration with \ac{nn} components.}

In general, \ac{mle} solutions can be understood as point estimates of Bayesian posteriors, assuming a \textit{uniform} prior~\citep[Ch.4]{pml1Book}.
When multiple parameter values explain the observations equally well, this simplification can result in ill-posed problems---causing \ac{mle} inverse games to recover potentially inaccurate estimates~\citep{li2023cost}. %
Moreover, in the context of motion planning, the use of point estimates without awareness of uncertainty can result in unsafe plans~\citep{peters2024ral,hu2022active}.

To address these issues, several works take a Bayesian view on inverse games~\citep{le2021lucidgames,peters2020accommodating}, aiming to infer a posterior \emph{distribution} on hidden game parameters while factoring in prior knowledge.
Since exact Bayesian inference is intractable in these problems, the belief update may be approximated via a particle filter \citep{peters2020accommodating}.
However, this approach requires solving a large number of equilibrium problems online to maintain the belief distribution, posing a significant computational burden.
A sigma-point approximation~\citep{le2021lucidgames} reduces the number of required samples but limits the estimator to unimodal uncertainty models.

\revised{
To obtain multi-hypothesis predictions tractably, \citet{diehl2023energy} and \citet{lidard2023nashformer} integrate game-theoretic layers in \acp{nn} for motion forecasting.
Both methods show that the inductive bias of games improves performance on real-world human datasets.
A game solver is used to refine predictions of a transformer model in~\citep{lidard2023nashformer}. 
\citet{diehl2023energy} embeds the solution of potential games~\citep{monderer1996potential} in a \ac{nn} to predict trajectory candidates. 
However, both approaches are limited to the prediction of a fixed number of trajectories
and offer no clear Bayesian interpretation of the learned model.
}

To overcome the limitations of  \ac{mle} approaches while avoiding the intractability of exact Bayesian inference \revised{over continuous game parameter distributions}, we propose to approximate the posterior via a \ac{vae}~\citep{kingma2013auto} that embeds a differentiable game solver~\citep{liu2023learning} during training.
The proposed approach can be trained from an unlabeled dataset of observed interactions, naturally handles continuous, multi-modal parameter and trajectory distributions, and does not require computation of game solutions at runtime to sample parameters from the posterior. 
Furthermore, to exploit crucial but subtle contextual information from visual cues when trajectory observations are uninformative or unavailable, we extend the proposed framework to jointly consume multi-modal observations.

\subsection{Multi-Modal \acp{vae}}

In the machine learning literature, numerous multi-modal \ac{vae} frameworks have been developed to fuse information across modalities (\eg vision, text, and audio), with particular emphasis on handling missing modalities and enabling cross-modal generation~\citep{schonfeld2019generalized,shi2019variational,wu2018multimodal,suzuki2016joint,sutter2021generalized}.

In robotics, multi-modal \acp{vae} are often used for sensor fusion, supporting perception, prediction, and control from heterogeneous inputs even when some modalities are unavailable~\citep{zhou2021vmloc,zambelli2020multimodal,meo2021multimodal}. For instance, \citet{zhou2021vmloc} use a variational product-of-experts formulation to fuse RGB and depth images for localization. \citet{zambelli2020multimodal} learn a shared sensorimotor latent space for humanoid control from state, vision, tactile, and sound. \citet{meo2021multimodal} incorporate a multi-modal \ac{vae} into an active inference controller to reactively control industrial manipulators from raw images and proprioceptive state estimates.

In autonomous driving, multi-modal \acp{vae} are used predominantly for motion forecasting~\citep{lee2017desire,salzmann2020trajectron++,yuan2021agentformer}, typically conditioning on agents’ past states together with contextual cues such as scene or semantic maps. Unlike multi-modal \ac{vae} formulations that explicitly enforce a shared latent space across modalities~\citep{schonfeld2019generalized}, these forecasting models generally treat multi-modal inputs as conditioning variables rather than aligning them through a joint latent representation. Learning a shared latent space~\citep{schonfeld2019generalized} is instead most beneficial when modalities may be missing at test time (e.g., when new agents enter the scene, we may have only camera imagery but have not observed a history of their behavior).

More broadly, physics-informed \acp{vae} incorporate differentiable physical structure---via simulators, constraint layers, or physics-based priors/dynamics---to ground (part of) the latent space in physically meaningful quantities, while retaining standard amortized variational inference~\citep{xu2022physicsVAE,ali2023physics,takeishi2021physics}. Along these lines, our work embeds a differentiable Nash game solver into the \ac{vae}, grounding sampling in interpretable game parameters while learning a joint latent space across modalities that remains effective when agents’ history states are partially missing.

\section{Preliminaries}\label{sec:preliminaries}
This paper studies interactions of rational, self-interested agents through the lens of parameterized \emph{\acfp{gnep}}. 
Within this setup, each of the $N$ agents aims to independently minimize their individual cost, fully aware that the remaining agents---their \say{opponents}---are similarly pursuing their own \emph{individual} objectives.
We formulate a parametric $N$-agent \ac{gnep} through $N$ interconnected constrained optimization problems, expressed as
\begin{subequations}
\label{eq:forward-gnep}
\begin{align}
\mathcal{S}^i_\gameParams(\traj^{\neg i}) \defeq
\arg\min_{\traj_i}\quad  & J^i_\gameParams(\traj^i, \traj^{\neg i})\\
\mathrm{s.t.}
\quad & g^i_\gameParams(\traj^i, \traj^{\neg i}) \geq 0,
\end{align}
\end{subequations}
where $\gameParams \in \reals^p$ parameterize agents' objectives and constraints. For each agent $i\in[N]\defeq\{1,\ldots,N\}$,  $J^i_\gameParams$ denotes their cost function and $g^i_\gameParams$ denotes their private constraints.
Notably, both the cost and constraints for agent~$i$ are sensitive not only to their own strategy $\traj^i \in \reals^{m_i}$ but also to the collective strategies of the other agents, $\traj^{\neg i} \in \reals^{\sum_{j\in[N]\setminus\{i\}}{m_j}}$.
Throughout this paper, we use the terms \emph{agent} and \emph{player} interchangeably to refer to one of the $N$ decision-making entities in the game.

\smallskip\noindent\textbf{Generalized Nash Equilibria.}
Given parameters $\gameParams$, the \emph{solution} to a \ac{gnep} is a~\ac{gne}: a \emph{strategy profile} $\mathbf{\traj}^* \defeq (\traj^{1*}, \ldots, \traj^{N*})$ in which every agent's strategy represents an optimal response to the strategies of all others, meaning
\begin{align}\label{eq:gne-condition}
   \traj^{i*} \in \mathcal{S}_\gameParams^i(\traj^{\neg i*}), \quad \forall i \in [N].
\end{align}
In essence, at a \ac{gne}, no agent can unilaterally reduce their cost by deviating to a different feasible strategy.

\smallskip\noindent\textbf{Example: Game-Theoretic Motion Planning.}
The aforementioned game-theoretic framework can naturally be used for online trajectory planning for a robot interacting with other \emph{uncontrolled} agents (humans or other robots).
\Cref{fig:motivation} exemplifies such a scenario.
Here, each agent~$i$'s strategy $\traj_i$ corresponds to a \emph{trajectory} (a finite-horizon sequence of states and controls).
From a practitioner's perspective, the procedure for setting up \emph{game-theoretic} motion planning is similar to that for setting up standard \emph{optimization-based} motion planning: the practitioner writes code that, for each agent $i$, specifies constraints~$g_\gameParams^i(\cdot)$ that encode state/input limits, dynamics, and obstacle avoidance, and an objective~$J_\gameParams^i(\cdot)$ that captures the $i^\mathrm{th}$ agent's preferences, such as making progress towards a goal location.
Typically, both constraints and objectives will depend on the current \emph{state} of the robot and its opponents (e.g., to adapt to changing lane boundaries or speed limits, and avoid collisions).
During online execution, the robot invokes a game solver to find an equilibrium solution to this trajectory game in a receding-horizon manner, resulting in a decision-making paradigm analogous to \ac{mpc}: \emph{\ac{mpgp}}.
In this context, the equilibrium solution of the game serves two purposes: with the convention that index~1 denotes the \emph{ego} agent, the opponent trajectories in the equilibrium solution, $\traj^{\neg1*}$, provide a game-theoretic \emph{prediction} of opponents' future decisions, and the ego agent's trajectory, $\traj^{1*}$, yields the robot's corresponding best response.

\begin{remark}
The structure of~\cref{eq:gne-condition}---in which each strategy $\traj^{i*}$ depends on the strategies of all other players $\traj^{\neg i*}$, and vice versa---underscores the \emph{interdependence} of agents' strategies in \acp{gnep}.
A player cannot simply predict others' strategies and optimize its own in response, as altering its own strategy may, in turn, affect those predictions. Thus, solving an \ac{gnep} entails reasoning about both prediction and planning \emph{simultaneously}.
This characteristic makes \acp{gnep} powerful for modeling tightly-coupled, interdependent decision-making.
\end{remark}

\smallskip\noindent\textbf{The Role of Game Parameters $\boldsymbol{\gameParams}$.}
A primary distinction between multi-agent trajectory games and single-agent trajectory optimization lies in the need to specify the costs and constraints for \emph{every} agent involved.
In real-world scenarios, a robot often lacks full information about its opponents' objectives, dynamics, or states, making it challenging to construct a fully specified game-theoretic model.
This aspect motivates the \emph{parameterized} game formulation introduced earlier: moving forward in this work, $\gameParams$ will represent the uncertain elements of the game.
In the setting of game-theoretic motion planning, $\gameParams$ commonly encompasses features of opponents' cost functions and constraints, such as their unspecified target lane or desired speed.
To streamline notation, we represent game~(\ref{eq:forward-gnep}) succinctly as the parametric collection of problem components $\game(\gameParams) \defeq(\{J^i_\gameParams, g^i_\gameParams\}_{i \in [N]})$.
In the following sections, we discuss methods for estimating these parameters online based on observed behaviors.

\section{Approach}\label{sec:approach}
When a robot interacts with others, it typically does not know their intents a priori.
Therefore, to interact safely and efficiently, a robot must \emph{infer} the intents of others from observed player behavior online.
Our primary contribution addresses exactly this problem through a game-theoretic lens.

The key idea underlying our approach is to invert the question posed in~\cref{sec:preliminaries}:
assuming that the observed behavior is the result of solving a game with unknown parameters, we seek to find those game parameters that explain the observed behavior.
For this reason, this process is commonly termed the \emph{inverse game} problem~\citep{waugh2013computational}.

Prior work on inverse games, such as that by \citet{peters2023ijrr}, \citet{li2023cost}, and \citet{clarke2023learning}, has primarily focused on inferring game parameters by maximizing the likelihood of observed behaviors. Although this \ac{mle} technique is widely used, it has notable limitations: (\romannumeral 1) it produces only a single \say{best-fit} estimate of $\gameParams$, making it difficult to capture or quantify uncertainty, which is critical for tasks like motion planning; and (\romannumeral 2) it can fail in scenarios where the available observations provide insufficient or ambiguous information, resulting in poor parameter recovery.
We demonstrate these limitations through experiments in \cref{sec:experiments}.

Furthermore, unlike prior work on inverse games, including the conference paper underpinning this work~\citep{liu2024auto}, the approach presented here incorporates not only low-dimensional state observations (such as positions and velocities) but also \emph{high-dimensional} sensor data such as images.
As we show in \cref{sec:experiments}, this novel capability to process visual cues within inverse games allows our method to reduce uncertainty about other players' intents faster than prior approaches, thereby leading to safer and more efficient interactions.

\subsection{A Bayesian View on Inverse Games}
\label{sec:bayesian-inverse-games}
We adopt a \emph{Bayesian} perspective on inverse games as follows.
Let $\observation=(\observedTrajectory, \observedImage)\in\observationSpace$ represent the \emph{joint observation} associated with the solution of a game parameterized by $\gameParams$, composed of two modalities: a trajectory observation~$\observedTrajectory$ and an image observation~$\observedImage$.

We aim to compute the posterior (belief) over the game parameters~$\gameParams$ given this multi-modal observation, \ie
\begin{align}\label{eq:bayesian-inverse-game}
    b(\gameParams) \defeq p(\gameParams \given \observation) %
\end{align}
Unlike the \ac{mle} formulation, which yields only a point estimate $\gameParams_{\mathrm{MLE}} \in \arg \max_{\gameParams} p(\observation \given \gameParams)$, this Bayesian approach infers the entire posterior over the unknown game parameters~$\gameParams$, and naturally incorporates prior information.

\subsubsection{Observation Modalities}\label{sec:approach-obs-modality}
Trajectory and image data have distinct characteristics, which we describe below.

\para{Trajectory Observations, $\observedTrajectory$}
Trajectory observations are low-dimensional, partial state observations over a fixed-lag window of the recent past (\eg in our running example, noisy position and velocity measurements over the last 10 time steps).
This modality gives the robot a sense of the physical state and recent actions taken by other players.
Taking an inverse game approach, we assume that trajectory observations are the direct consequence of all agents playing a game with unknown parameters.
Consistent with this assumption, like prior works~\citep{le2021lucidgames,li2023cost,peters2023ijrr}, we choose a Gaussian observation model for the trajectory observations, \ie
\begin{align}\label{eq:trajectory-game-observation-model}
    p(\observedTrajectory\given\gameParams) \defeq \normaldist(\underbrace{(\toTrajectoryObservationMean\circ\solveGame)(\gameParams)}_{\mu_{\observedTrajectory}(\gameParams)}, \Sigma_\observedTrajectory),
\end{align}
where $\solveGame$ denotes a game solver that maps game parameters $\gameParams$ to an equilibrium trajectory profile $\traj^*$ of the game $\game(\gameParams)$, and~$\toTrajectoryObservationMean$ denotes a function that maps that equilibrium trajectory profile $\traj^*$ to the mean of the partial trajectory observation distribution (\eg extracting position and velocity from the trajectory).
For simplicity, we assume that the covariance matrix $\Sigma_\observedTrajectory$ is fixed and known.

\para{Image Observations, $\observedImage$}
Image observations provide additional visual cues about other players' intents.
Unlike trajectory observations, image observations cannot easily be related to the game solution alone via a simple Gaussian observation model.
For example, in the intersection scenario in~\cref{fig:motivation}, the same trajectory profile~$\traj^*$ may be observed with vastly different images: the scene may play out under different weather conditions, times of day, between different types and colors of cars, and on different road geometries.
Yet, driving behavior and image observations are correlated in subtle ways: \eg sports cars may be more likely to drive aggressively, trucks may be less likely to make sudden turns and tend to stay in the right lane, and drivers may be more cautious on a rainy day.

It is intractable to model these subtle correlations manually via a simple, user-specified observation model.
Instead, we will learn an \emph{implicit} observation model directly from an unlabeled dataset of joint observations~$\dataset = \{\observation_k = (\observedTrajectory_k, \observedImage_k)\mid \observation_k \sim p(\observation), \forall k \in [K]\}$.

\subsection{Auto-Encoding Bayesian Inverse Games}
In theory, the inverse game problem of~\cref{eq:bayesian-inverse-game} can be solved simply via Bayes' rule
$$
    p(\gameParams \given \observation) = \frac{p(\observation \given \gameParams) p(\gameParams)}{p(\observation)}.
$$
However, beyond the challenge that an observation model for complex sensor data (such as images) is typically not available, several challenges make direct inference of the posterior intractable:
\begin{enumerate}
    \item The prior $p(\gameParams)$ is typically unavailable and instead must be learned from data.
    \item The computation of the normalizing constant, $p(\observation) = \int p(\observation \given \gameParams) p(\gameParams) \mathrm{d}\gameParams$, is intractable in practice due to the marginalization of $\gameParams$.
    \item Both the prior $p(\gameParams)$ and the posterior $p(\gameParams \given \observation)$ are in general non-Gaussian or even \emph{multi-modal} and are therefore difficult to represent explicitly in terms of their \ac{pdf}.
\end{enumerate}
Prior work~\citep{le2021lucidgames} partially mitigates these challenges by using a \ac{ukf} for approximate Bayesian inference, but that approach is limited to unimodal uncertainty models, low-dimensional observations, and requires solving multiple games for a single belief update, thereby posing a computational challenge.

Fortunately, as we demonstrate in \cref{sec:experiments}, many practical applications of inverse games do not require an explicit evaluation of the belief \ac{pdf}, $b(\gameParams)$.
Instead, a \emph{generative model} of the belief---\ie one that allows us to draw samples $\gameParams \sim b(\gameParams)$---often suffices.
In this section, we demonstrate how to learn such a generative model from an unlabeled dataset $\dataset = \{\observation_k = (\observedTrajectory_k, \observedImage_k)\mid \observation_k \sim p(\observation), \forall k \in [K]\}$ of observed interactions by building a game-theoretic \acf{vae}.

\subsubsection{Introducing a Latent Variable Model}
\begin{figure}
    \centering
    \includegraphics[width=0.5\linewidth]{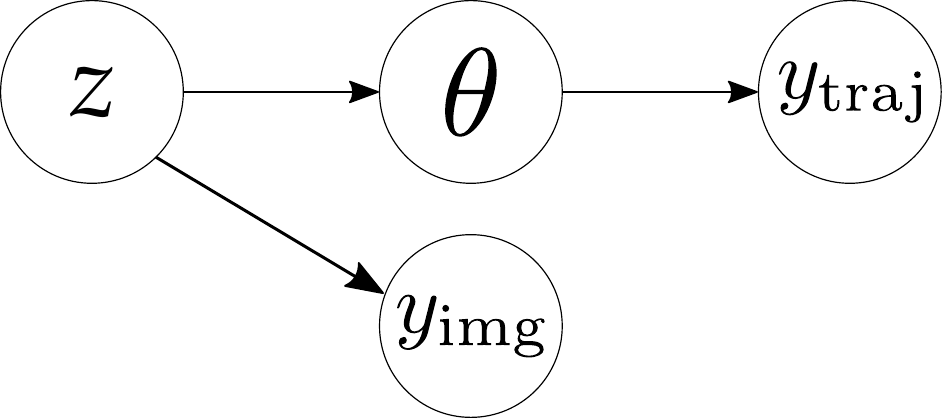}
    \caption{Bayes network representation of our latent variable model relating the observed image $\observedImage$ and trajectory $\observedTrajectory$ to the game parameters $\gameParams$ via a latent variable $\latent$. Note that, by construction, the observed image $\observedImage$ and trajectory $\observedTrajectory$ are statistically \textit{dependent} but \textit{conditionally independent}, given latent variable $\latent$.}
    \label{fig:latent-variable-model}
\end{figure}

To learn a generative model of the belief $b(\gameParams)$, we introduce a latent variable model as summarized in~\cref{fig:latent-variable-model}.
Following this model structure, in addition to the trajectory observation model $p(\observedTrajectory\given\gameParams)$ of~\cref{eq:trajectory-game-observation-model}, we define the following distributions:
\begin{subequations}
    \label{eq:latent-variable-model}
    \begin{align}
        p(\latent)                                    & \defeq \normaldist(0, I),                                          \\
        p_\decoderParams(\gameParams\given\latent)    & \defeq \delta(\gameParams - \gameDecoder(\latent)),                \\
        p_\decoderParams(\observedImage\given\latent) & \defeq \normaldist(\imageDecoder(\latent), \Sigma_\observedImage),
    \end{align}
\end{subequations}
where $\delta$ denotes the Dirac delta function,
$\imageDecoder$ and $\gameDecoder$ denote decoder \acp{nn} that map the latent variable $\latent\in\latentSpace$ to the game parameters~$\gameParams$ and the mean of the image observation~$\mu_\observedImage$, respectively,
and $\Sigma_\observedImage$ denotes the covariance matrix of the image observation.

Due to the deterministic relationship between $\gameParams$ and $\latent$, we can further express the trajectory observation model directly conditioned on $\latent$ as
\begin{align}\label{eq:trajectory-observation-model-conditioned-on-latent}
    p_\decoderParams(\observedTrajectory\given\latent) & \defeq \normaldist((\toTrajectoryObservationMean\circ\solveGame\circ\gameDecoder)(\latent), \Sigma_\observedTrajectory),
\end{align}
where $\Sigma_\observedTrajectory$ denotes the covariance matrix of the trajectory observation.

Naturally, the latent variable model is only useful when it explains the data.
To this end, we will seek a parameterization $\decoderParams^*$ such that the induced data distribution
\begin{align*}\label{eq:induced-data-distribution}
    p_{\decoderParams^*}(\overbrace{\observedTrajectory, \observedImage}^{\observation}) \defeq
    \int
    p_{\decoderParams^*}(\observedTrajectory\given\latent)
    p_{\decoderParams^*}(\observedImage\given\latent)
    p(\latent)
    \mathrm{d}\latent,
\end{align*}
matches the true data distribution $p(\observation)$, \ie $\dkl{p_\decoderParams(\observation)}{p(\observation)} \approx 0$, where
$$
    \dkl{p}{q} := \expectedValue{x\sim p(x)}{\log p(x) - \log q(x)}
$$
denotes the \ac{kl} divergence between arbitrary distributions $p$ and $q$.

\subsubsection{Properties of the Latent Variable Model}
\label{sec:properties-of-latent-variable-model}
The newly introduced latent variable~$\latent$ takes on no physical interpretation.
However, the chosen model structure has a useful property: it allows us to express the prior and posterior over game parameters as an inference problem over the latent variable~$\latent$.
Specifically,
we can express the prior over game parameters as
\begin{align}
    p_\decoderParams(\gameParams) & \defeq \int p_\decoderParams(\gameParams\given\latent)p(\latent)\mathrm{d}\latent,
\end{align}
and the posterior over game parameters as\footnote{
    In~\cref{eq:posterior-game-parameters}, the conditional independence of $\gameParams$ and $\observation$ given $\latent$ is due to the deterministic relationship between $\gameParams$ and $\latent$, which implies that $p_\decoderParams(\gameParams \given \observation,\latent) \equiv p_\decoderParams(\gameParams \given \latent)$.
}
\begin{align}\label{eq:posterior-game-parameters}
    p_\decoderParams(\gameParams\given\observation) & \defeq \int p_\decoderParams(\gameParams\given\latent)p_\decoderParams(\latent\given\observation)\mathrm{d}\latent.
\end{align}
This analysis reveals that we can generate samples from $p_\decoderParams(\gameParams)$ by sampling from $p(\latent)$ (which is known to be a multivariate Gaussian by construction) and passing the samples through the decoder network $\gameDecoder$.
Similarly, if we map samples from $p_\decoderParams(\latent\given\observation)$ through the decoder network $\gameDecoder$, we generate samples from $p_\decoderParams(\gameParams\given\observation)$.

However, sampling from the exact posterior $p_\decoderParams(\latent\given\observation)$ directly requires marginalizing over $\gameParams$ and $\latent$, and is generally intractable.
Therefore, we will fit a Gaussian surrogate model $q_\encoderParams(\latent\given\observation)$ such that $\dkl{q_\encoderParams(\latent\given\observation)}{p_\decoderParams(\latent\given\observation)} \approx 0$, as is common in amortized variational inference~\citep{murphy2023probabilistic}.
Since $\gameDecoder$ is highly non-linear, despite the Gaussian approximation in latent space, this will allow us to capture non-Gaussian, multi-modal distributions over game parameters~$\gameParams$.
We will demonstrate our method's capacity to capture such multi-modal distributions in \cref{sec:experiments}.

Before we discuss how to train the latent variable model~$p_\decoderParams$ and the surrogate model $q_\encoderParams$, let us briefly connect our approach to the terminology of \acp{vae}.

\subsubsection{Connection to Variational Autoencoders}\label{sec:amortized-inference}
In the terminology of \acp{vae}, the surrogate model $q_\encoderParams(\latent\given\observation)$ is known as an \emph{encoder}.
Furthermore, the distribution
\begin{align}\label{eq:structured-decoder}
    p_\decoderParams(\overbrace{\observedTrajectory, \observedImage}^{\observation}\given\latent) & \defeq
    p(\observedTrajectory\given\latent)p(\observedImage\given\latent)
\end{align}
corresponds to the \emph{decoder} in the terminology of \acp{vae}.

The key difference between our approach and a conventional \ac{vae} is the special \emph{structure} of this decoder.
While a conventional \ac{vae} employs an unstructured \ac{nn} as an observation model, our observation model contains the game solver $\solveGame$ as a special \say{layer}.
It is this game-theoretic \say{layer} in the decoding pipeline that induces an interpretable structure on the output of the decoder \ac{nn} $\gameDecoder$, forcing it to predict the hidden game parameters $\gameParams$. %
\Cref{fig:pipeline} illustrates this interpretation of our approach as a structured \ac{vae}.

\subsection{Training a Game-Theory-Informed Variational Autoencoder}\label{sec:training-game-theory-informed-vae}

Below, we outline the process for optimizing the model parameters of the latent variable model~$p_\decoderParams$ and the encoder $q_\encoderParams$ jointly.
Having established the connection to conventional \acp{vae}, we begin by reviewing the high-level training process, which closely matches that of the literature \citep{kingma2013auto,murphy2023probabilistic} in~\cref{sec:high-level-training-process}.
Next,~\cref{sec:game-specific-training} highlights special considerations due to the embedded game solver $\solveGame$.

\subsubsection{High-Level Training Process}\label{sec:high-level-training-process}
The parameters of the latent variable model~$p_\decoderParams$ and the encoder $q_\encoderParams$ can be optimized jointly by maximizing the \ac{elbo} over the data distribution $p(\observation)$.
Specifically, we can express the log-likelihood of the joint observation~$\observation$ as
\begin{subequations}
    \begin{align}
         & \log p_\decoderParams(\observation) \nonumber                                                                                                                                                          \\
         & = \log \int p_\decoderParams(\observation\given\latent)p(\latent)\mathrm{d}\latent                                                                                                                     \\
         & = \log \int \frac{p_\decoderParams(\observation\given\latent)p(\latent)}{q_\encoderParams(\latent\given\observation)}q_\encoderParams(\latent\given\observation)\mathrm{d}\latent                      \\
         & = \log \expectedValue{z\sim q_\encoderParams(\latent\given\observation)}{\frac{p_\decoderParams(\observation\given\latent)p(\latent)}{q_\encoderParams(\latent\given\observation)}}                    \\
         & \geq \expectedValue{z\sim q_\encoderParams(\latent\given\observation)}{\log p_\decoderParams(\observation\given\latent) + \log p(\latent) - \log q_\encoderParams(\latent\given\observation)}\nonumber \\
         & =: \ell(\decoderParams, \encoderParams, \observation).
    \end{align}
\end{subequations}
Hence, in order to fit the parameters $\decoderParams$ and $\encoderParams$, we can maximize\footnote{
    It is easy to verify that maximizers of this objective also maximize the log-likelihood of the data while minimizing the KL divergence between the true posterior and the surrogate posterior, since $\ell$ can be equivalently written as \vspace{-1em}
    $$
        \ell(\decoderParams, \encoderParams, \observation) =
        \log p_\decoderParams(\observation) - \dkl{q_\encoderParams(\latent\given\observation)}{p_\decoderParams(\latent\given\observation)},
        \vspace{-1em}
    $$
    cf. \cite{murphy2023probabilistic}.

}
\begin{align}\label{eq:high-level-optimization-problem}
    \decoderParams^*, \encoderParams^* \in \arg\max_{\decoderParams, \encoderParams} \expectedValue{\observation\sim p(\observation)}{\ell(\decoderParams, \encoderParams, \observation)}.
\end{align}
In practice, this optimization problem is solved via stochastic gradient ascent over (batched) samples from the dataset~$\dataset$, replacing the outer expectation with a sample average.
This requires propagating gradients through the inner expectation over $q_\encoderParams(\latent\given\observation)$ involving the observation model $p_\decoderParams(\observation\given\latent)$.

Like conventional \acp{vae}, we can use the ``reparameterization trick''~\citep{kingma2013auto} to compute the gradient of the inner expectation.
Even with this trick, however, computation of gradients~$\nabla_{\decoderParams, \encoderParams} \ell$ faces an additional challenge: gradient propagation through the decoder~$p_\decoderParams(\observation\given\latent)$ in \cref{eq:structured-decoder} requires propagating gradients through the trajectory observation model $p_\decoderParams(\observedTrajectory\given\latent)$, which in turn requires propagating gradients through the game solver $\solveGame$ in \cref{eq:trajectory-observation-model-conditioned-on-latent}.
We will discuss how to address this challenge in the next section.

\subsubsection{Gradient Propagation Through the Game-Theory-Informed Observation Model}\label{sec:game-specific-training}
To optimize~\cref{eq:high-level-optimization-problem}, we need to compute gradients of the decoder~$p_\decoderParams(\observation\given\latent)$ with respect to both the decoder parameters~$\decoderParams$ and latent samples~$\latent$.
While large parts of this process can be performed via automatic differentiation, expanding the chain rule on~\cref{eq:trajectory-observation-model-conditioned-on-latent} reveals that both operations involve backpropagation through the game solver $\solveGame$.
To make this backpropagation tractable, we apply the implicit differentiation approach proposed in our previous work~\citep{liu2023learning}.
We include a brief summary of that approach here for completeness.

\smallskip\noindent\textbf{Local Equilibrium Conditions.}
To facilitate gradient propagation through the game solver, we analyze the local equilibrium conditions of the game.
For a game $\game(\gameParams)$ as defined in~\cref{eq:forward-gnep}, these local equilibrium conditions correspond to the stacked \acf{kkt} conditions for all players:
\begin{align}\label{eq:local-equilibrium-conditions}
    \forall i \in [N]: \left\{ \begin{aligned}
                                   \nabla_{\traj^i} \lagrangian^i(\traj^i, \traj^{\neg i}, \dualMultiplier^i, \gameParams) & = 0,    \\
                                   0 \leq g^i_\gameParams(\traj^i, \traj^{\neg i}) \perp \dualMultiplier^i                 & \geq 0,
                               \end{aligned} \right.
\end{align}
where $\lagrangian^i$ denotes the \emph{Lagrangian} for player~$i$,
\begin{align*}
    \lagrangian^i(\traj^i, \traj^{\neg i}, \dualMultiplier^i, \gameParams) \defeq J^i_\gameParams(\traj^i, \traj^{\neg i}, \gameParams) - \dualMultiplier^{i\transpose} g^i_\gameParams(\traj^i, \traj^{\neg i}, \gameParams).
\end{align*}
We employ the PATH solver~\citep{dirkse1995path} to find primals $\traj^{i*}$ and duals $\dualMultiplier^{i*}$ that satisfy these conditions.

\smallskip\noindent\textbf{Implicit Differentiation.}
The \ac{kkt} conditions provide an implicit relationship between the primal solution~$\traj^{i*}$, the duals $\dualMultiplier^{i*}$, and the game parameters $\gameParams$.
We exploit this relationship to compute the sensitivity of the solution~$\traj^{i*}$ with respect to the game parameters $\gameParams$ as follows.
Let $\activeIndices^i$ denote the rows of $g^i_\gameParams(\traj^{i*}, \traj^{\neg i*}, \gameParams)$ that hold with equality at the solution,
let $\neg\activeIndices^i$ denote the complement of that set,
and let $\extractRows{\activeIndices}{\cdot}$ denote the operator that returns the rows associated with the indices in $\activeIndices$.
Using this notation, we define $\activeConditions$ to extract all conditions that hold with equality at the solution,
\begin{align*}
    \activeConditions(\traj^*,\dualMultiplier^*,\gameParams) & \defeq
    \begin{bmatrix}
        \vdots                                                                                  \\
        \nabla_{\traj^i} \lagrangian^i(\traj^i, \traj^{\neg i}, \dualMultiplier^i, \gameParams) \\
        \extractRows{\activeIndices^i}{g^i_\gameParams(\traj^i, \traj^{\neg i}, \gameParams)}   \\
        \extractRows{\neg\activeIndices^i}{\dualMultiplier^i}                                   \\
        \vdots
    \end{bmatrix},
\end{align*}
stacked for all players $i~\in~[N]$.
Additionally, we define $\primalsAndDuals := (\traj^1, \dualMultiplier^1, \dots, \traj^N, \dualMultiplier^N)$ to be the stacked vector of all players' primals and duals.
Assuming that none of the constraints are weakly active~(in that both the primal constraint holds with equality and the associated dual variable takes the value zero), small changes in $\gameParams$ must be compensated by changes in the primals and duals such that $\activeConditions$ remains zero.
Mathematically, this means that the \emph{total derivative} of $\activeConditions$ with respect to the game parameters $\gameParams$ must be zero.
Expanding this total derivative reveals
\begin{align}\label{eq:total-derivative-of-active-conditions}
    \frac{\mathrm{d}\activeConditions}{\mathrm{d}\gameParams} =
    \frac{\partial\activeConditions}{\partial\gameParams} +
    \frac{\partial\activeConditions}{\partial\primalsAndDuals} \frac{\partial\primalsAndDuals}{\partial\gameParams} = 0,
\end{align}
which provides us with a linear system of equations that we can solve for $\frac{\partial\primalsAndDuals}{\partial\gameParams}$, revealing the partial derivatives of the game solution with respect to the game parameters.\footnote{
    Certain edge cases are not discussed here. For example, the solution may include weakly active constraints, or \cref{eq:total-derivative-of-active-conditions} may not admit a unique solution if $\frac{\partial\primalsAndDuals}{\partial\gameParams}$ is rank-deficient. We refer the reader to our previous work~\citep{liu2023learning} for details on how to handle these edge cases.
}

\begin{remark}
    Note that the derivative computation in~\cref{eq:total-derivative-of-active-conditions} has low computational overhead compared to the forward pass of the game solver: since, in the forward pass, we already compute $\primalsAndDuals$, the only additional computation is the partial differentiation of the active conditions with respect to $\gameParams$ and $\primalsAndDuals$ to set up the linear system of equations~\cref{eq:total-derivative-of-active-conditions}.
    Our open-source implementation~\citepalias{MCPTrajectoryGameSolver} automates this procedure while tightly integrating with automatic differentiation frameworks in~Julia.
\end{remark}

\begin{figure*}
    \centering
    \includegraphics[width=0.75\linewidth]{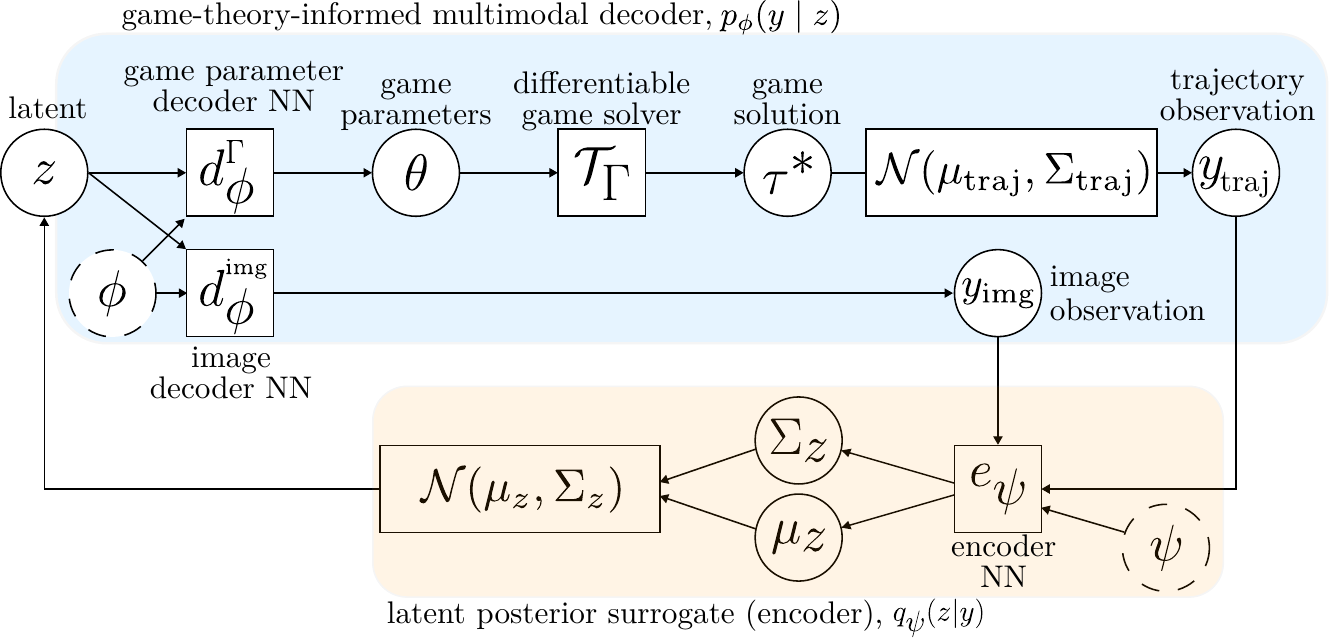}
    \caption{Overview of a structured \ac{vae} for generative Bayesian inverse games.}
    \label{fig:pipeline}
\end{figure*}

\subsubsection{Remark on Inference Time.} At inference time, to generate samples from the estimated posterior $q_{\encoderParams,\decoderParams}(\gameParams \given \observation) \defeq \int p_\decoderParams(\gameParams \given \latent)q_\encoderParams(\latent\given\observation) \mathrm{d}\latent$, we do not need to evaluate the game solver $\solveGame$ or the image decoder~$\imageDecoder$: posterior sampling involves only the evaluation of the \acp{nn} $\gameDecoder$ and $\encoder$; cf.\ $\observation\to\gameParams$ in \cref{fig:pipeline}.
As a result, given a joint observation~$\observation$ of images and trajectories, we can run the encoder \emph{once} to obtain the latent distribution $q_\encoderParams(\latent\given\observation)$ and then repeatedly sample from this distribution to recover a large number of posterior samples from $q_{\encoderParams,\decoderParams}(\gameParams \given \observation)$ at low computational cost.
Consequently, this inference pipeline can operate repeatedly at high frequencies, enabling the generation of updated posteriors in real time for downstream receding-horizon motion planning.

\section{Experiments}\label{sec:experiments}

This section presents empirical results that test the following hypotheses about the proposed Bayesian inverse game framework:

\begin{itemize}
    \item \textbf{(H1, Inference quality)}: The proposed framework infers unknown objectives more accurately than \ac{mle}-based inverse game methods, especially in settings where multiple objectives can explain the same observations. 
    \item \textbf{(H2, Multi-modal uncertainty awareness)}: The framework quantifies uncertainty in objective inference and captures multimodal structure in both the learned prior and posterior distributions.
    \item \textbf{(H3, Planning safety)}: Using the inferred posterior in downstream planning yields safer robot behavior than planning based on an \ac{mle} inverse game solution.
    \item \textbf{(H4, Multi-modal observations)}: The framework supports integrating additional observation modalities (e.g., visual cues alongside partial-state trajectories) for objective inference.
    \item \textbf{(H5, Limited/uninformative trajectory history)}: When trajectory history is limited (e.g., an agent has just entered the scene) or uninformative, fusing multimodal observations improves inference of unknown opponent intent and reduces posterior uncertainty.
    \item \textbf{(H6, Safety and comfort)}: The reduced uncertainty and earlier intent inference enabled by multimodal observations enhance planning safety and improve motion comfort by avoiding unnecessary steering.
\end{itemize}

\subsection{Experiment Setup and Baselines}

In our main evaluation experiments (\cref{sec:intersection-state-obs,sec:multi-modal-obs}), we consider a two-agent intersection scenario (cf. \cref{fig:motivation}) in which an ego robot interacts with an opponent whose intent is unknown to the ego. Unless otherwise specified, the opponent’s ground-truth intent is drawn from an equally weighted two-component Gaussian mixture, with modes corresponding to turning left or proceeding straight through the intersection.

As an auxiliary study, \cref{sec:highway} evaluates a simplified highway scenario to more closely and qualitatively illustrate the posterior distributions inferred by the proposed approach. The highway setup is described in \cref{sec:highway}.

To simulate strategic interaction, we generate the opponent’s actions by solving trajectory games in a receding-horizon fashion using the agents’ ground-truth intents. The ego agent is controlled by solving the same receding-horizon trajectory games, but using the opponent intent inferred by the evaluated methods detailed in \cref{sec:baselines}.

We model each agent $i$ using a kinematic bicycle model with state $x_t^i = (p_{x,t}^i, p_{y,t}^i, v_t^i, \xi_t^i)$ and control $u_t^i = (a_t^i, \eta_t^i)$ at time step $t$, where $(p_{x,t}^i, p_{y,t}^i)$ denotes position, $v_t^i$ the longitudinal velocity, $\xi_t^i$ the heading, $a_t^i$ the acceleration, and $\eta_t^i$ the steering angle. In \cref{sec:intersection-state-obs,}, agents are controlled directly by decentralized, receding-horizon game solutions. In \cref{sec:multi-modal-obs}, agents are controlled by low-level PID controllers in the CARLA simulator~\citep{Dosovitskiy17} that track the corresponding decentralized, receding-horizon game solutions.

For clarity, we index the ego agent as $i=1$ and the opponent as $i=2$. At each time step, each agent minimizes its cost over a planning horizon of $T=15$:
\begin{align}\label{eq:game-cost}
J^i_\theta 
= {} & \sum_{t=1}^{T-1} 
\|p_{t+1}^i - p_{\mathrm{goal}}^i\|_2^2
+ 0.1 \|u_t^i\|_2^2 \notag \\
& + 400 \max\!\left(
  0,\, d_{\min} - \|p_{t+1}^i - p_{t+1}^{\neg i}\|_2
\right)^3,
\end{align}
where $p_{t}^i = (p_{x,t}^i, p_{y,t}^i)$ denotes agent $i$'s position at time $t$, $p_{\mathrm{goal}}^i$ is agent $i$'s goal position, $d_{\min}$ is the minimum allowable inter-agent distance, and $\theta$ denotes the opponent’s latent intent. This objective captures the trade-off among goal reaching, control effort, and collision avoidance. In this example, the unknown opponent intent parameter $\theta$ corresponds to the opponent’s two-dimensional goal position. 

In \cref{sec:intersection-state-obs}, the ego robot receives partial-state observations of the opponent, consisting of its position and orientation in the past $15$ time steps. In \cref{sec:multi-modal-obs}, we additionally evaluate a variant of our approach that incorporates a raw image frame of the opponent for inference. We extract $768$-dimensional visual features from the raw images using a pretrained DINOv3 ViT-B/16 model~\citep{oquab2024dinov3}, and pass these features to the structured \ac{vae}.

Both the encoder $\encoder$ and decoder $\decoder$ of the structured \ac{vae} in \cref{sec:approach} are implemented as fully connected feedforward networks with two hidden layers. In \cref{sec:intersection-state-obs}, we use a $16$-dimensional latent variable $\latent$ and hidden widths of $128$ and $80$ for the encoder and decoder, respectively. In \cref{sec:multi-modal-obs}, to accommodate the higher-dimensional observations, we use a $64$-dimensional latent variable $\latent$, with hidden widths of $512$ and $320$ for the encoder and trajectory decoder, respectively, and a hidden width of $512$ for the image decoder.

We train the structured \ac{vae} using interaction data generated from closed-loop game-play, without labels of the players’ intents. Specifically, we simulate interactions by repeatedly solving the ground-truth game in a receding-horizon fashion, with the opponent’s true goal $\theta$ sampled from the two-mode Gaussian mixture described above. From each interaction episode, we construct training samples by sliding a length-$15$ time-step window along the trajectory to form observation sequences containing the agents’ positions and orientations.

For the multi-modal structured \ac{vae} in \cref{sec:multi-modal-obs}, we additionally pair each trajectory observation sequence with a top-down RGB image frame captured at the end of the window, reflecting the opponent’s most recent configuration. In \cref{sec:intersection-state-obs} we use $560$ interaction episodes for training; in \cref{sec:multi-modal-obs}, we use $700$ episodes to account for the increased complexity of that setting. We train each structured \ac{vae} with Adam~\citep{kingma2014adam} for approximately 14 hours of wall-clock time on a 32-core desktop.

\subsubsection{Baselines}\label{sec:baselines}

We evaluate our Bayesian inverse game framework in motion planning tasks where an autonomous robot interacts with an opponent agent whose intent is unknown. 
At each step, the ego robot receives new observations of the opponent and infers the opponent’s intent. Given this inverse game solution, the ego robot then applies standard game-theoretic motion planning methods by solving the corresponding games parameterized by the inferred objectives to compute its receding-horizon actions. We compare the following method combinations: 

\noindent\textbf{(i) Ground truth (GT)}. This method serves as an \say{oracle} with access to the opponent agent’s ground truth intention.

\noindent\textbf{(ii) Bayesian inverse game (ours) + \ac{bpine}}. This method first solves Bayesian inverse games using the proposed approach, then plans for the robot in a contingency 
framework~\citep{peters2024ral} by minimizing the expected cost
\(
\expectedValue{\costparams \sim q_{\encoderParams,\decoderParams}(\gameParams \given \observation)}{J_{\costparams}^1},
\)
where the expectation is taken under the inferred posterior over objectives.

\noindent\textbf{(iii) Bayesian inverse game (ours) + \acl{map} planning (B-MAP)}. This method first solves Bayesian inverse games using the proposed approach, then extracts a \ac{map} estimate
\(
\hat{\costparams}_{\mathrm{MAP}} \in \arg\max_{\gameParams} q_{\encoderParams,\decoderParams}(\gameParams \given \observation),
\)
and solves the corresponding game $\game(\hat{\gameParams}_{\mathrm{MAP}})$.

\noindent\textbf{(iv) Randomly initialized \ac{mle} planning (R-MLE)}. This method solves an \ac{mle} inverse game,
\(
\hat{\costparams}_{\mathrm{MLE}} \in \arg\max_{\gameParams} p(\observation \given \gameParams),
\)
using the online gradient-descent approach of \citet{liu2023learning}, and then solves the game $\game(\hat{\gameParams}_{\mathrm{MLE}})$ under the same game structure as in our methods above. The initial guess for the online inverse game optimization is sampled uniformly from a rectangular region that covers the candidate ground-truth goals.

\noindent\textbf{(v) Bayesian prior initialized \ac{mle} planning (BP-MLE)}. This method is identical to R-MLE, except that the \ac{mle} inverse game optimization is initialized by sampling the opponent intention from the learned Bayesian \emph{prior} of our approach, rather than from a uniform distribution.

\noindent\textbf{(vi) Static Bayesian prior planning (St-BP)}. This method samples opponent intentions from the learned Bayesian prior and directly solves a game parameterized by the sampled objective. It serves as an ablation to isolate the effect of performing inverse game objective inference on downstream planning performance.

For methods that use the proposed \ac{vae} for inverse game inference, we draw 1000 samples at each time step to approximate the posterior distribution. Thanks to amortized inference, this sampling procedure runs in real time and takes approximately \SI{7}{ms}, avoiding the heavy online optimization required by prior inverse game approaches~\citep{liu2023learning,peters2023ijrr}. For B-PinE, we cluster the posterior samples into two groups to match the multi-hypothesis game solver in~\cite{peters2024ral}. We note that, for more complex belief distributions, modern computational methods exist to parallelize the computation~\citep{li2023scenario}.

\subsection{Intersection Scenario with Partial-State-Only Observations}
\label{sec:intersection-state-obs}

\begin{figure*}
  \centering
  \includegraphics[width=\linewidth]{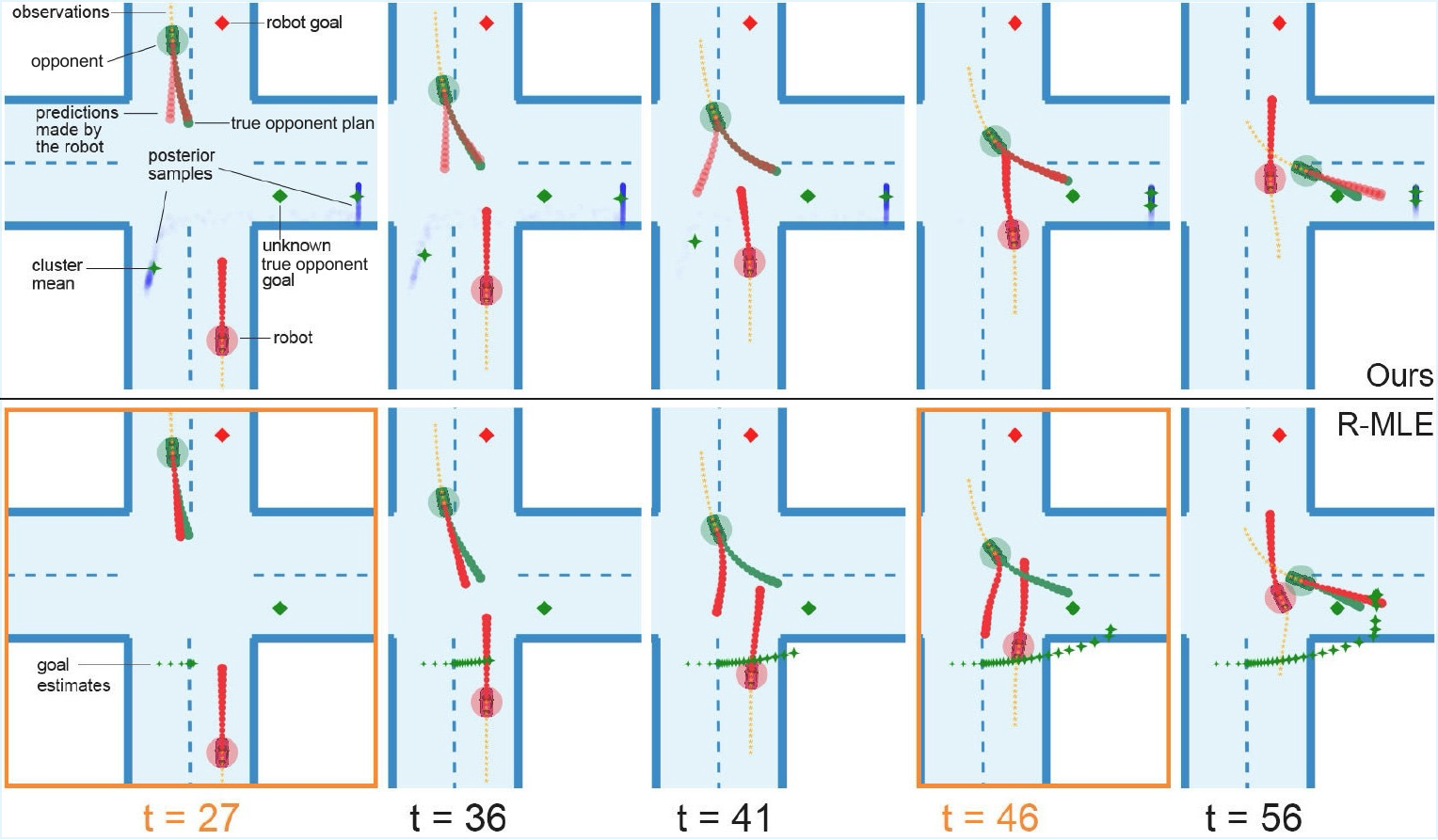}
  \caption{Snapshots from an intersection scenario (\cref{sec:intersection-state-obs}) in which the ego robot is controlled by our \ac{bpine} approach (top) and the R-\ac{mle} baseline (bottom). For R-\ac{mle}, the estimated goal is shown as a green star whose size increases over time. Reproduced from conference publication~\citep{liu2024auto}.
  }
  \label{fig:intersection-qualitative}
\end{figure*}

\begin{figure}
  \centering
  \includegraphics[width=0.9\linewidth]{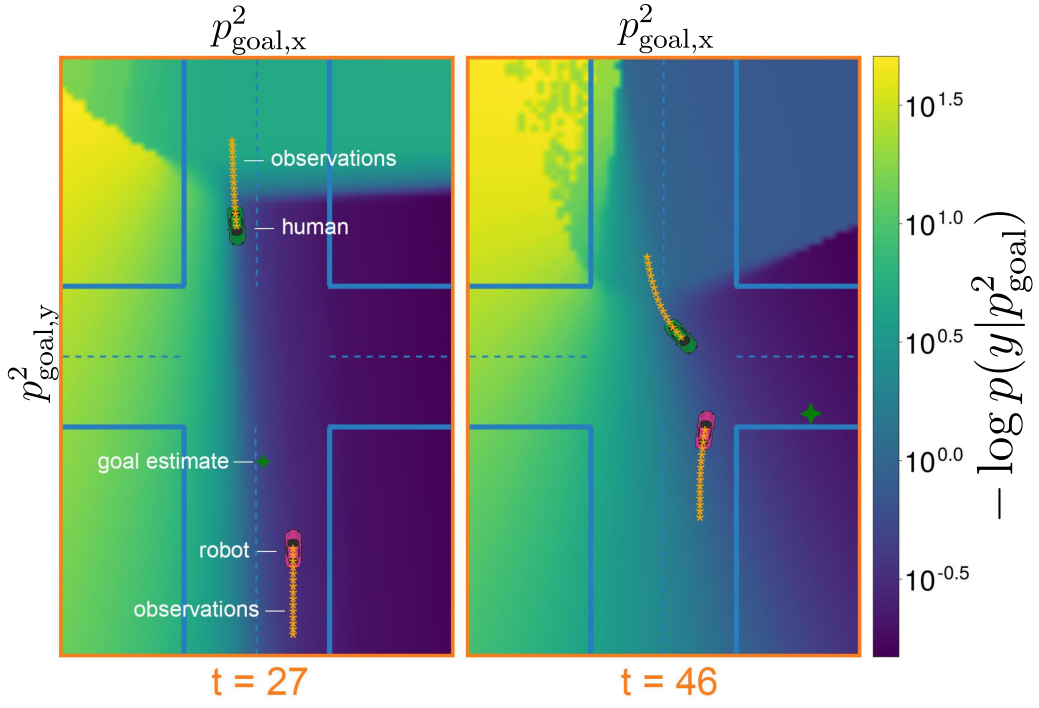}
  \caption{
  \ac{mle} inverse game cost landscape (negative observation log-likelihood, $-\log p(\observation \given p^2_{\mathrm{goal}})$, over opponent goal positions $p^2_{\mathrm{goal}}$) for the R-\ac{mle} baseline at time steps 27 and 46 of the interaction in \cref{fig:intersection-qualitative}. Because it ignores Bayesian priors, the \ac{mle} inverse game \emph{problem} can be ill-posed and exhibit a flat cost landscape; consequently, its solutions may induce unsafe downstream decisions (cf. \cref{fig:intersection-qualitative}). Reproduced from conference publication~\citep{liu2024auto}.
  }
  \label{fig:mle-cost}
\end{figure}

We first evaluate the proposed approach in a simulated intersection-driving scenario with only trajectory observations. \Cref{fig:intersection-qualitative} shows snapshots of representative behaviors produced by our \ac{bpine} approach (top) and the R-\ac{mle} baseline (bottom).

For \ac{bpine}, before the opponent enters the intersection, the observed state history is insufficient to uniquely identify the opponent’s intent—whether it will go straight or turn left. Accordingly, our method infers a bimodal posterior in which both intents are roughly equally likely. The downstream game-theoretic planner therefore produces a multimodal prediction of the opponent’s future motion and plans conservatively to account for both possibilities: the robot yields and passes the opponent on the left. As the opponent approaches the intersection and its behavior becomes more informative, the posterior collapses to a unimodal distribution, the intent is fully disambiguated, and the robot successfully resolves the interaction.

In contrast, before the opponent enters the intersection, the R-\ac{mle} baseline returns a point estimate that trivially concludes the opponent intends to go straight—perfectly explaining the observed behavior while \emph{ignoring} prior knowledge that an alternative intent may also be plausible. Lacking uncertainty awareness, the robot (incorrectly) commits with high confidence to the straight-going hypothesis and plans to traverse the intersection aggressively. When the opponent’s true intent is eventually revealed, the \ac{mle} solution shifts toward the correct objective; however, the robot has already committed to an aggressive maneuver and can no longer adjust in time to avoid a collision.

\subsubsection{Observability challenge of \ac{mle} inverse games}

To better understand the failure mode of the R-\ac{mle} baseline in \cref{fig:intersection-qualitative}, \cref{fig:mle-cost} zooms in on the underlying \ac{mle} inverse game problems solved online by R-\ac{mle} and visualizes the resulting \ac{mle} cost landscape, \ie the negative observation log-likelihood $-\log p(\observation \given p^2_{\mathrm{goal}})$ over opponent goal positions $p^2_{\mathrm{goal}}$, at two representative time steps. Recall that the \ac{mle} formulation simplifies Bayesian inference by ignoring the prior and greedily maximizing the observation likelihood:
\[
p\!\left(p^2_{\mathrm{goal}} \given \observation\right) \propto
\underbrace{p\!\left(\observation \given p^2_{\mathrm{goal}}\right) p(y)}_{\mathrm{Bayesian}}
 \textcolor{red}{\cancel{\propto}}
\underbrace{p\!\left(\observation \given p^2_{\mathrm{goal}}\right)\textcolor{red}{\xcancel{p(y)}}}_{\mathrm{\ac{mle}}}.
\]
As a consequence, before the opponent enters the intersection and its motion becomes informative of its intent, the \ac{mle} cost landscape can be nearly flat (cf.\ the dark regions in \cref{fig:mle-cost}). In this regime, many goal hypotheses explain the observations equally well, so the \ac{mle} solver may return an essentially arbitrary point estimate, which can lead to unsafe downstream decisions. In contrast, our Bayesian inverse-game approach explicitly incorporates the prior and quantifies posterior uncertainty, enabling the robot to plan conservatively when the opponent's intent is ambiguous.

\emph{Overall, these qualitative behaviors support hypothesis H1 and the posterior-focused component of H2.}

\subsubsection{Monte Carlo Evaluation}

\begin{figure*}
  \centering
  \includegraphics[width=0.8\linewidth]{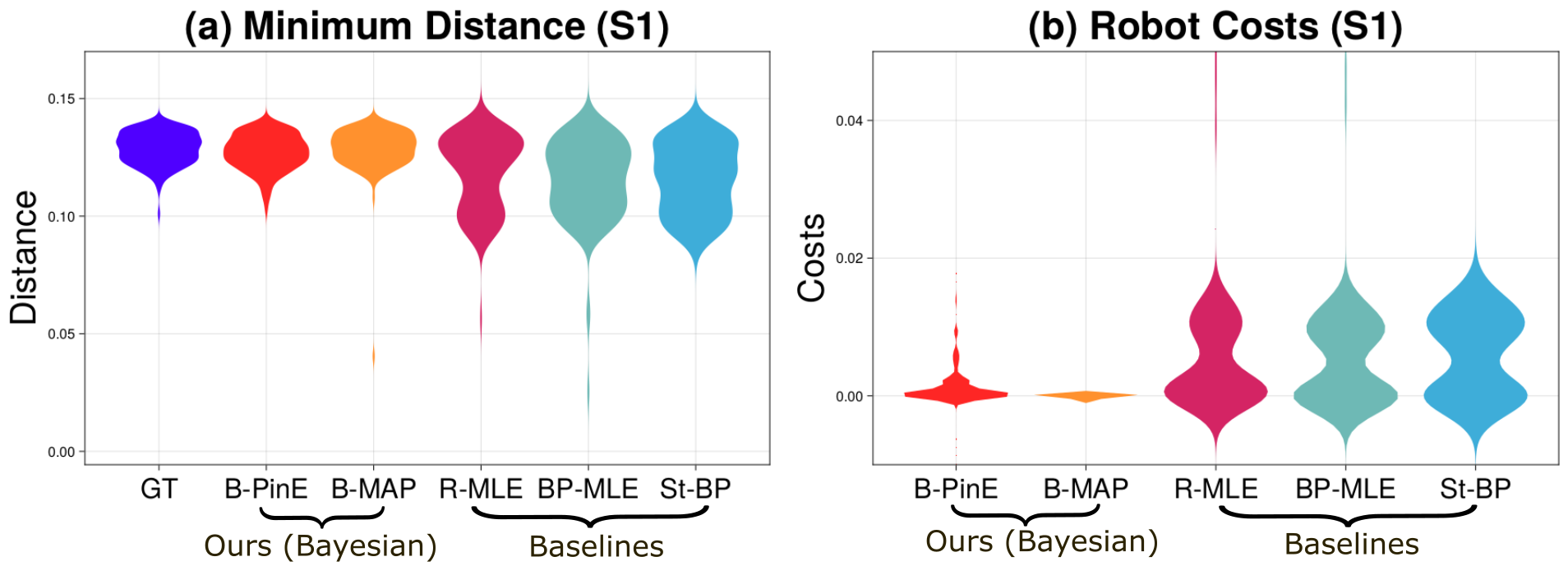}
  \caption{
Monte Carlo study results of the intersection scenario in \cref{fig:intersection-qualitative} (\cref{sec:intersection-state-obs}) for trials where the ego robot passes after the left-turning opponent. Bayesian inverse game methods provide significantly better safety (higher minimum inter-agent distance) and efficiency (lower ego robot cost) than \ac{mle}-based inverse game methods and the baseline that does not solve inverse games. Robot costs are reported after subtracting the ground-truth game costs. Reproduced from conference publication~\citep{liu2024auto}.
  }
  \label{fig:intersection-quantitative-ego-second}
\end{figure*}
\begin{figure*}
  \centering
  \includegraphics[width=\linewidth]{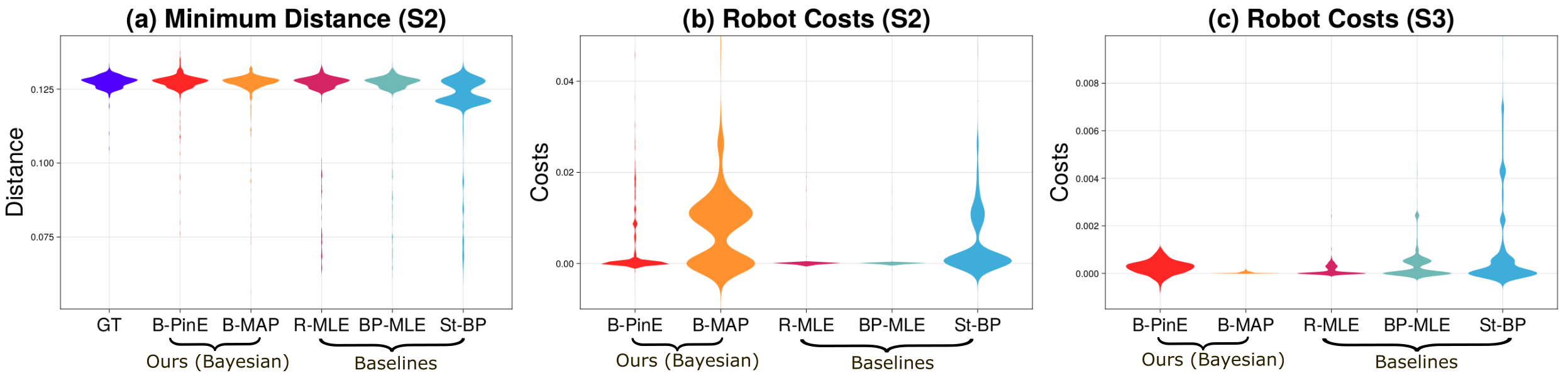}
  \caption{
Monte Carlo study results for the intersection scenario in \cref{fig:intersection-qualitative} (\cref{sec:intersection-state-obs}): (a–b) correspond to trials where the ego robot enters the intersection before a left-turning opponent, and (c) corresponds to trials where the opponent drives straight without interacting with the ego robot. Reproduced from conference publication~\citep{liu2024auto}.
  }
  \label{fig:intersection-quantitative-ego-first}
\end{figure*}

We quantitatively evaluate the methods in~\cref{sec:baselines} via a Monte Carlo study with 1500 simulation trials of the intersection scenario in \cref{fig:intersection-qualitative}. In each trial, the ego robot’s initial position is sampled uniformly along its lane, producing a diverse set of initial conditions and interactions—ranging from cases where the ego robot enters the intersection first to cases where the opponent arrives first. We report two metrics: (i) the minimum inter-agent distance over the rollout as an indicator of planning safety, and (ii) the ego robot’s incurred cost as a measure of efficiency.

We partition the 1500 trials into three categories based on the opponent’s goal and the initial conditions: \textbf{(S1)} the opponent turns left and reaches the intersection before the ego robot. This safety-critical setting features an aggressive opponent; the ego robot must promptly infer the opponent’s intent and yield to ensure safety. \textbf{(S2)} the opponent turns left, but the ego robot reaches the intersection first and can pass comfortably. Here, accurate intent identification is less critical because the ego robot has a timing advantage.
\textbf{(S3)} the opponent proceeds straight through the intersection. Safety is easy to satisfy, but the ego robot should ideally infer this intent early to avoid overly conservative behavior that degrades efficiency.

\Cref{fig:intersection-quantitative-ego-second} reports the Monte Carlo results for trials in \textbf{(S1)}. 
Methods that use our Bayesian inverse game solution achieve \emph{significantly better planning safety (higher minimum inter-agent distance) and efficiency (lower ego robot cost)}.
In this scenario, solving \ac{mle} inverse games roughly matches the performance of St-BP, \ie directly planning with our Bayesian priors without solving any inverse game. This result highlights the importance of accurate knowledge of the prior distribution. Taking the minimum inter-agent distance over all ground-truth trials as the collision threshold, the collision rates are \textbf{0.0\% for B-PinE}, \textbf{0.78\% for B-MAP}, 17.05\% for R-MLE, 16.28\% for BP-MLE, and 17.83\% for St-BP. Between B-PinE and B-MAP, B-PinE plans with the full posterior and is therefore more conservative (better safety) at the cost of slightly higher robot cost.

\Cref{fig:intersection-quantitative-ego-first} (a–b) shows the Monte Carlo results for trials in \textbf{(S2)}. As expected, the difference in safety performance between our Bayesian methods and the \ac{mle}-based baselines is less pronounced in this setting, since the ego agent enters the intersection first. Nonetheless, our Bayesian methods still improve safety: using the same collision distance threshold as in \textbf{(S1)}, the collision rates are \textbf{0.86\% for B-PinE}, \textbf{2.24\% for B-MAP}, 7.59\% for R-MLE, 6.03\% for BP-MLE, and 7.59\% for St-BP.

In terms of planning efficiency, B-MAP relies only on point estimates from the full Bayesian posterior and commits to overconfident, aggressive maneuvers that cause coordination failures between the agents (\eg both attempt to enter the intersection simultaneously and then must brake). In contrast, the B-PinE variant of our approach exploits the full posterior for uncertainty-aware planning and achieves better efficiency (and safety), underscoring the importance of planning under the full posterior distribution. 

Finally, \cref{fig:intersection-quantitative-ego-first} (c) shows that in \textbf{(S3)}, B-MAP attains the highest planning efficiency, while B-PinE is, by construction, more cautious and produces a few slightly more conservative trials compared to the point-estimate-based methods. The non–inverse-game baseline St-BP clearly underperforms in this setting.

Taken together, these results demonstrate that \emph{our Bayesian inverse game approach enables safer downstream motion planning than the baselines and supports hypothesis H3.} Between the two Bayesian variants, B-PinE and B-MAP, we observe improved planning safety when using the full Bayesian posterior in \textbf{(S1–S2)} and improved efficiency in \textbf{(S2)}.

\subsection{Highway Scenario with Partial-State-Only Observations}\label{sec:highway}

\Cref{sec:intersection-state-obs} evaluates the proposed Bayesian inverse game framework in conjunction with downstream motion planning methods, testing the benefit of solving the full Bayesian problem for the overall planning pipeline. This section then considers a simplified scenario and further zooms in on the qualitative behavior of the inferred distributions produced by our approach, in isolation from the planning task.

We consider a highway driving scenario in which two agents travel in sequence on a single-lane road. Each agent has a desired speed, but while the front agent can follow its goal speed freely, the rear agent is responsible for avoiding collisions---if it wishes to drive faster than the front agent, it must slow down, making its true intended speed unobservable.

The two agents’ behaviors are simulated jointly by solving games using the ground truth goal speeds of both agents. 
Our approach observes both agents' trajectories and infers the rear agent's desired speed, assuming the front agent's desired speed is known. 
By construction, we expect the proposed Bayesian approach to produce a high-uncertainty posterior in cases where the rear agent’s true desired speed is unobservable because it is blocked by the front agent.

In this simplified highway scenario, each agent is modeled as a double integrator: the longitudinal position and velocity form the state, and acceleration is the control input. To infer the opponent’s latent intent, we use a \ac{vae} with the same encoder/decoder hidden-layer sizes as in \cref{sec:intersection-state-obs}, but with a \emph{one-dimensional} latent variable. Observations consist of 15 steps of both agents' velocities.

\begin{figure}[h]
    \centering
    \begin{subfigure}{0.45\textwidth}
        \centering
        \includegraphics[width=\linewidth]{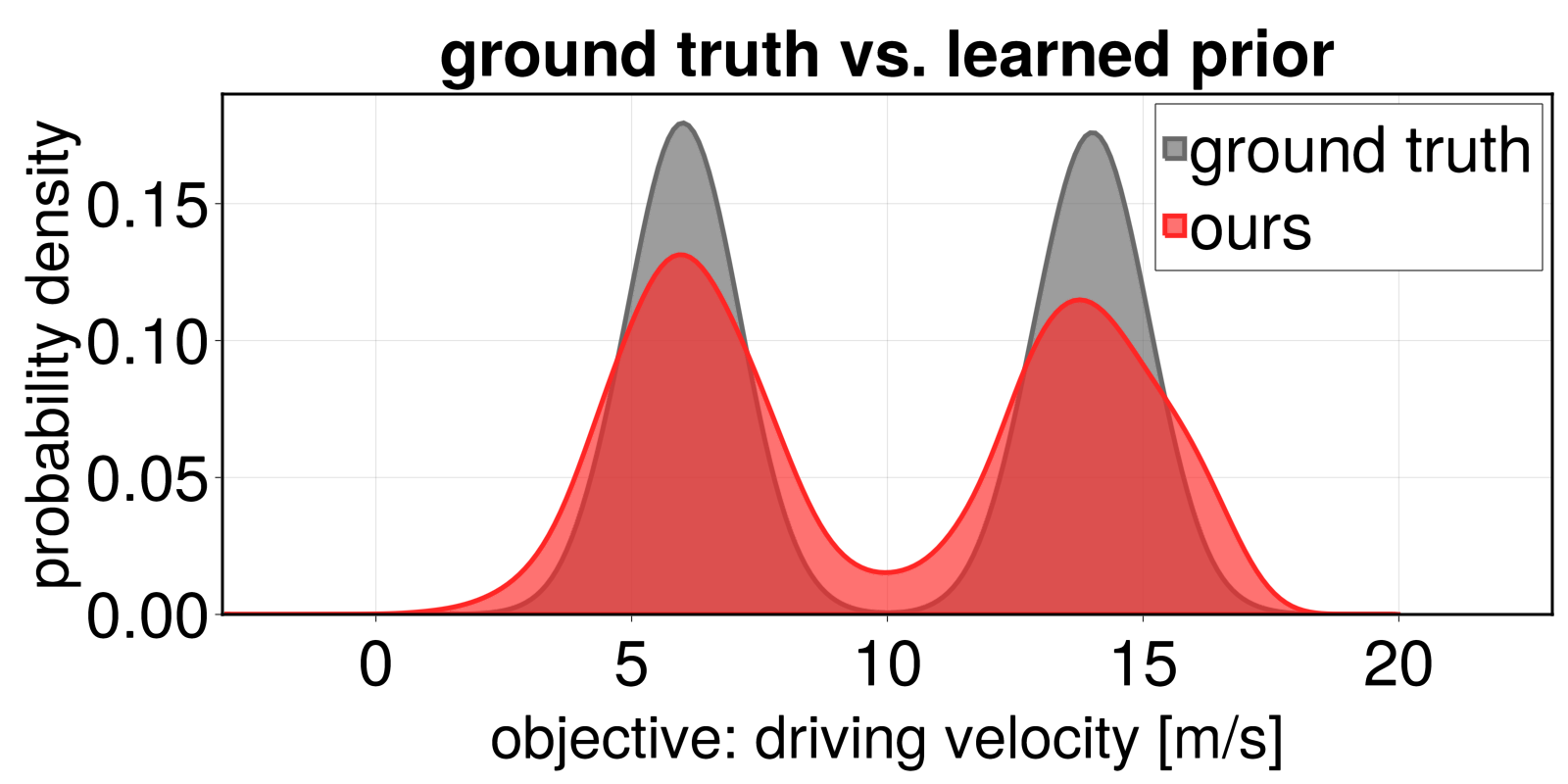}
        \caption{Prior distribution obtained by decoding prior latent samples.}
        \label{fig:highway-prior}
    \end{subfigure}

    \vspace{1em}

    \begin{subfigure}{0.5\textwidth}
        \centering
        \includegraphics[width=\linewidth]{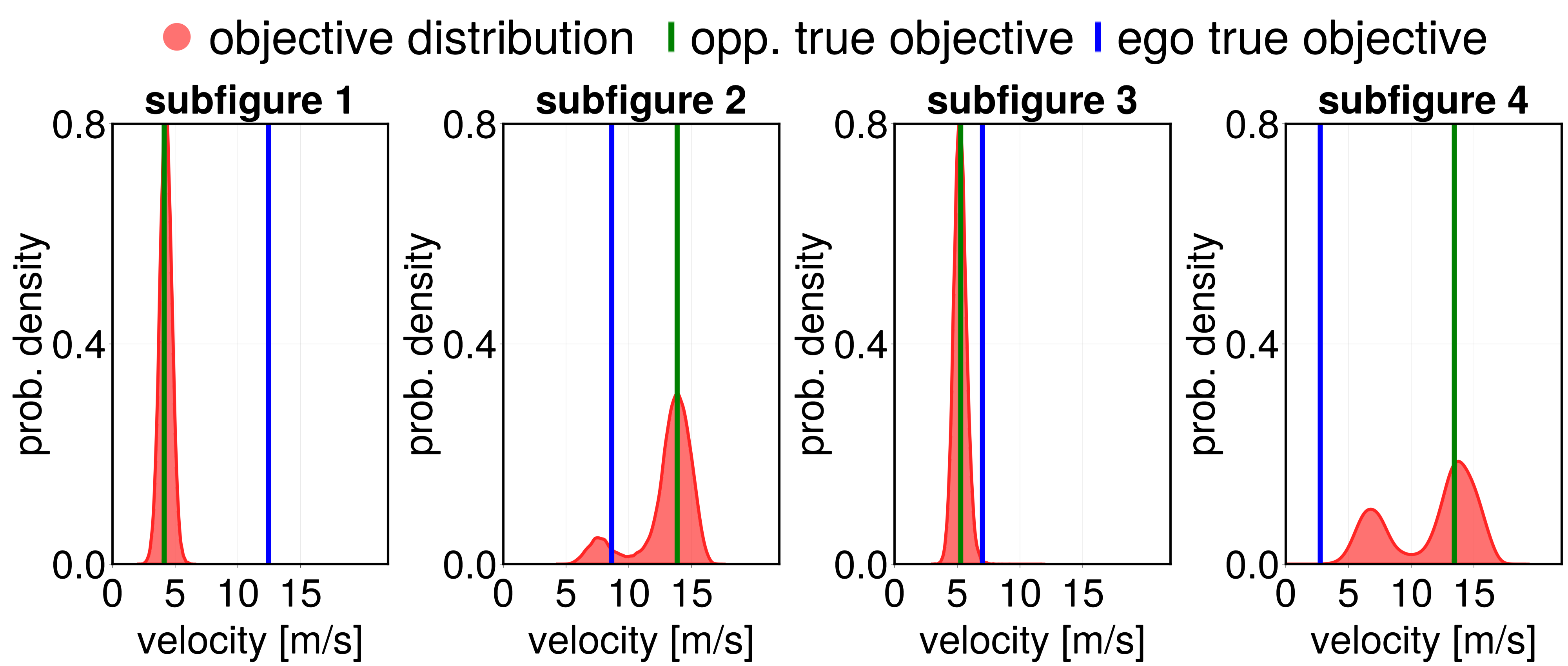}
        \caption{Posterior distribution obtained by decoding posterior latent samples.}
        \label{fig:highway-posterior}
    \end{subfigure}

    \caption{
Prior and posterior distributions produced by our Bayesian inverse game approach in a highway driving scenario (\cref{sec:highway}). The method successfully learns the prior from unlabeled data and exhibits uncertainty awareness in the posteriors: it produces a sharp distribution when the hidden parameter is identifiable from the observations, and a broad, multimodal distribution close to the prior when the observations are uninformative. Reproduced from conference publication~\citep{liu2024auto}.
    }
    \label{fig:highway-results}
\end{figure}

We train a structured \ac{vae} model on a dataset of 20{,}000 simulated observations. In each trial, the ego agent’s goal velocity is sampled uniformly from $\SI{0}{\meter\per\second}$ to $\SI{20}{\meter\per\second}$, while the opponent’s desired velocity is sampled from a bimodal Gaussian mixture (grey in \cref{fig:highway-results}, top) with two unit-variance components centered at 30\% and 70\% of the maximum velocity.

\Cref{fig:highway-prior} shows the learned prior distribution from a dataset of agents’ trajectories without labels for the rear agent’s intent. The prior is obtained by decoding samples from the prior latent space. The proposed Bayesian framework successfully recovers a bimodal prior distribution that closely matches the ground truth, whose importance was highlighted in previous experiments; cf. \cref{fig:mle-cost}.

\Cref{fig:highway-posterior} shows representative snapshots of the inferred posterior distribution. Whenever the front agent’s desired speed is higher than that of the rear agent, our approach produces a sharp, unimodal posterior that correctly captures the rear agent’s ground-truth intent with low uncertainty. In contrast, when the front agent blocks the rear agent—\ie when the rear agent’s desired speed is higher but it is constrained by the front agent—our approach produces a broad, bimodal posterior close to the prior, signaling high uncertainty.

Taken together with the results in \cref{sec:intersection-state-obs}, \emph{these findings support hypothesis H2: our Bayesian inverse game framework is uncertainty-aware and can capture multimodality in both the prior and posterior distributions}.

\subsection{Intersection Scenario with Multi-Modal Observations}\label{sec:multi-modal-obs}

\begin{figure*}
  \centering
  \includegraphics[width=\linewidth]{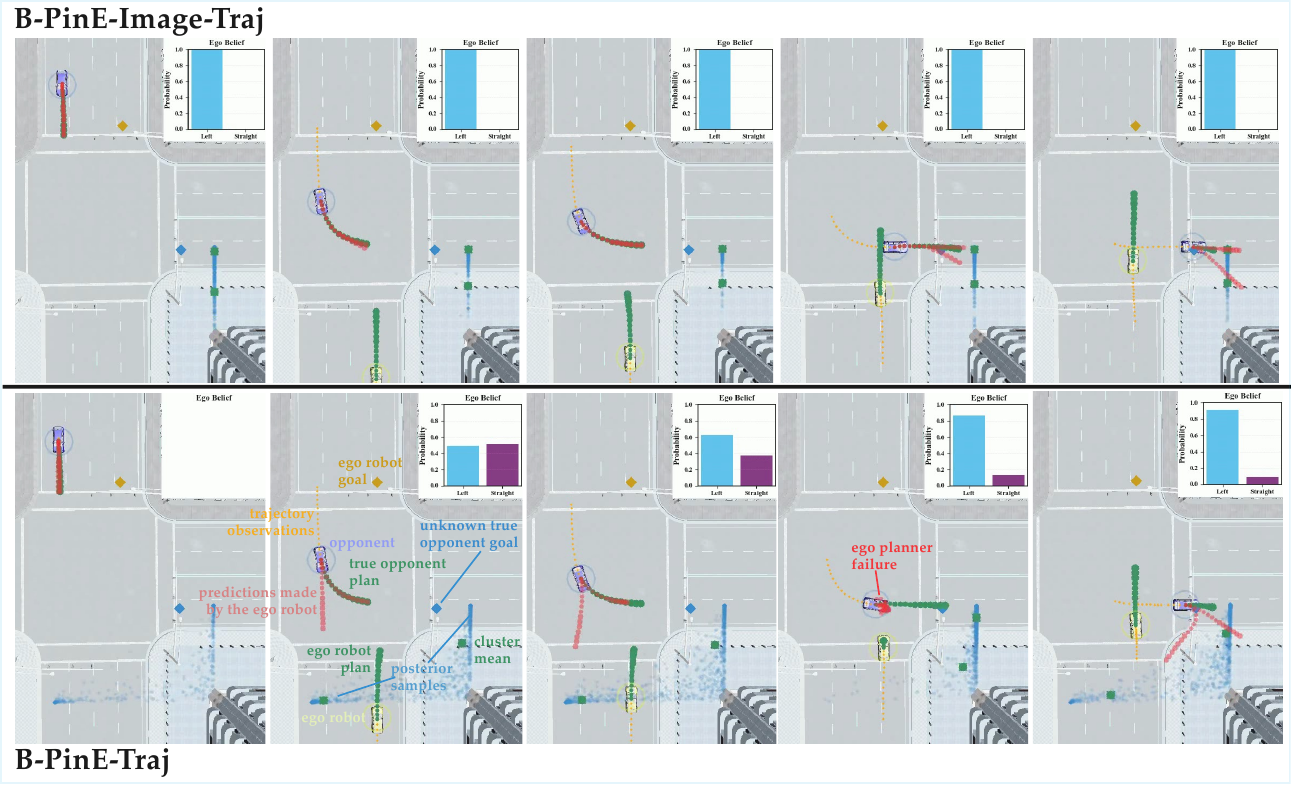}
  \caption{
Snapshots from an intersection scenario (\cref{sec:color-img}) in the CARLA simulator, where the opponent’s intent is encoded by its color. The ego agent controlled by B-PinE with an image-trajectory \ac{vae} (top) leverages visual contextual information to identify the opponent’s left-turning intent early and safely yields to resolve the interaction. In contrast, B-PinE with a trajectory-only \ac{vae} (bottom), shown in \cref{fig:intersection-qualitative}, recognizes the opponent’s intent too late, leading to a planner failure and an unsafe emergency hard brake.
}

  \label{fig:color-qualitative}
\end{figure*}

This section extends the evaluation of our Bayesian inverse game framework to incorporate multiple observation modalities per the discussion in \cref{sec:approach-obs-modality}, inferring unknown game parameters from joint image and partial-state trajectory observations. 

To evaluate the approach with photorealistic visual inputs, we study two variants—an image-trajectory \ac{vae} and a trajectory-only \ac{vae}—in the CARLA simulator~\citep{Dosovitskiy17}. In these experiments, vehicles are actuated by PID controllers that track decentralized, receding-horizon game solutions.

We evaluate two examples in the intersection scenario of \cref{fig:intersection-qualitative}, using datasets where image observations provide different levels of contextual information about the opponent’s intent—either through their color (\cref{sec:color-img}) or their vehicle type (\cref{sec:type-img}). In \cref{sec:color-img}, images are highly informative and effectively determine the modality of the opponent’s hidden goal (going straight or turning left), whereas in \cref{sec:type-img}, images provide only partial information that modestly complements the trajectory observations.

\subsubsection{Color-Encoded Agent Intents}
\label{sec:color-img}

This section evaluates the proposed framework in a setting where the opponent’s color provides information about its unknown intent. We construct a dataset in which, with probability $50\%$, the opponent is a blue car that turns left and, with probability $50\%$, the opponent is a red car that proceeds straight, as illustrated in \cref{fig:color-qualitative}. This setup mimics real-world scenarios in which an opponent explicitly signals intent, \eg via turn signals. We then train two instances of our structured \ac{vae}: one that uses both an image of the opponent and partial-state observations of their past behavior (as described above), and another that uses only partial-state observations.

In the image-trajectory \ac{vae}, image observations provide additional context for posterior inference. When the image contains information beyond what is available from the trajectory (as in this example), the posterior inferred from both modalities should be less uncertain and more closely aligned with the opponent’s ground-truth intent than the posterior inferred from trajectory data alone.

\paragraph{Qualitative Behavior.} We use the same B-PinE planner as in \cref{sec:intersection-state-obs} and compare closed-loop performance when coupled with either of the two structured \acp{vae} described above.
\Cref{fig:color-qualitative} illustrates the qualitatively different behaviors that can result from these two variants. When the opponent first enters the scene—and no trajectory history is yet available—the image-trajectory \ac{vae} successfully leverages strong prior information (the opponent’s color) from the image alone, together with trajectory statistics learned during training, to compute a narrow, accurate posterior over the opponent’s unknown goal (frame 1). In contrast, the trajectory-only \ac{vae} lacks sufficient information to form a posterior at this stage. As the interaction unfolds and more trajectory data become available, the image-trajectory \ac{vae} maintains its correct inference that the opponent is turning left and prepares the ego vehicle to yield. The trajectory-only \ac{vae}, however, first produces a bimodal posterior—remaining uncertain about whether the opponent will go straight or turn left—and only later collapses to the left-turn mode, by which time the B-PinE planner has already committed to aggressively passing on the right (frames 2-4). Consequently, the B-PinE planner with the image-trajectory \ac{vae} yields safely and resolves the interaction smoothly, whereas the B-PinE planner with the trajectory-only \ac{vae} gets too close to the oncoming vehicle and the game-theoretic planner is unable to find a feasible solution; ultimately, the ego vehicle executes an emergent hard brake.

\begin{figure*}
  \centering

  \begin{subfigure}{0.6\textwidth}
    \centering
    \includegraphics[width=\linewidth]{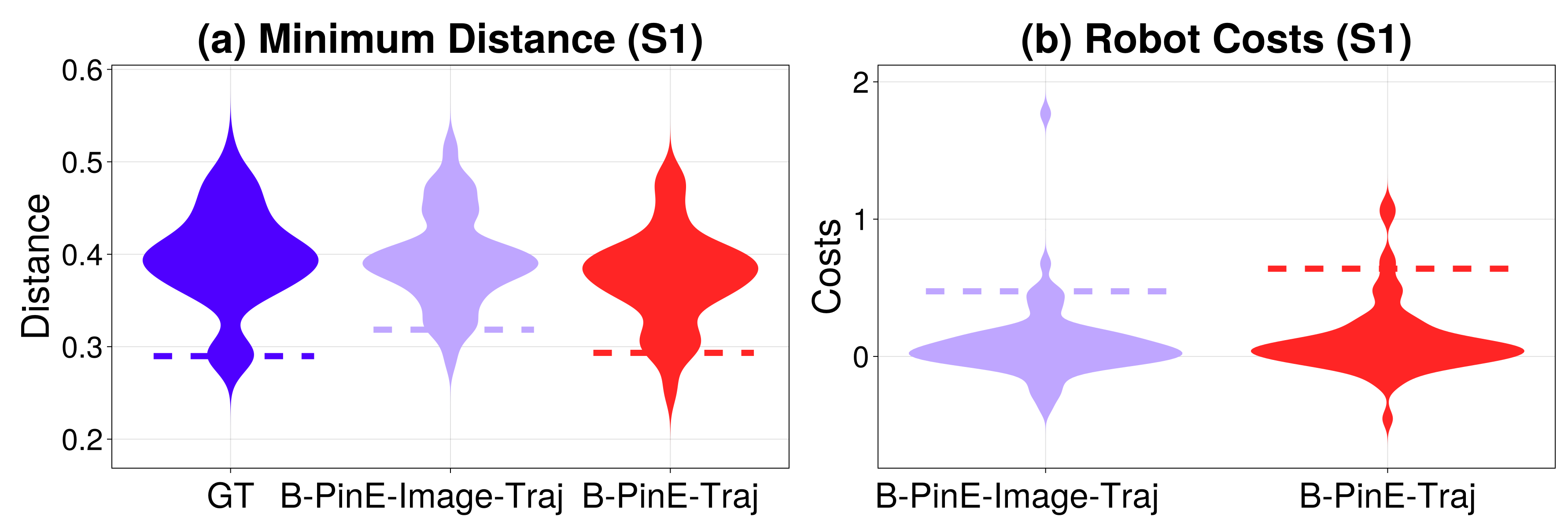}
    \caption{
Trials in which the ego robot passes through the intersection after a left-turning opponent.}
    \label{fig:fig-10b.png}

  \end{subfigure}

  \begin{subfigure}{\textwidth}
    \centering
    \includegraphics[width=\linewidth]{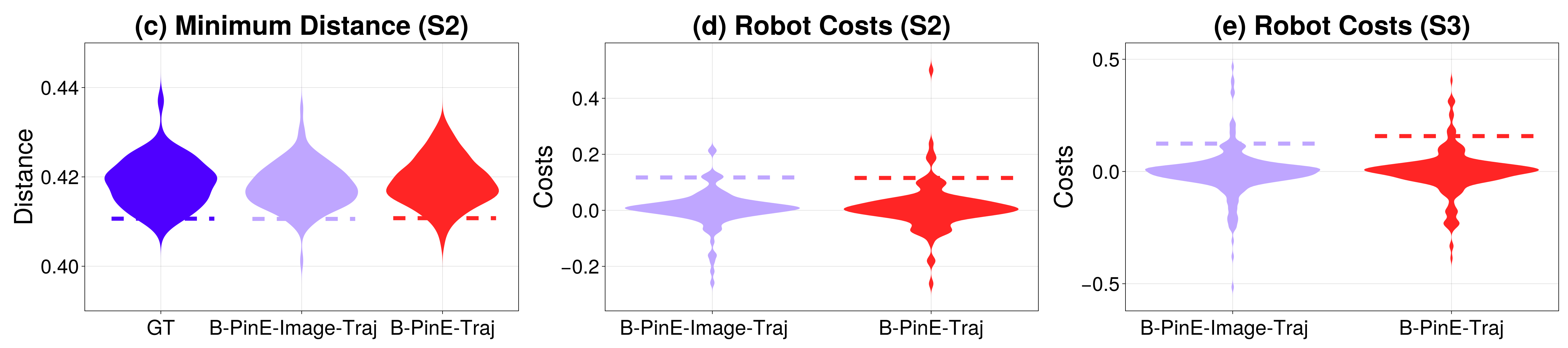}
    \caption{Trials in which the opponent turns left and the ego passes first (c–d), or the opponent goes straight (e).}
    \label{fig:color-ego-second}
  \end{subfigure}

  \vspace{0.8em} %

  \caption{
Monte Carlo results for the intersection scenario in \cref{fig:color-qualitative} (\cref{sec:color-img}). Panels (a, c) report minimum inter-agent distances for trials \textbf{(S1--2)}, and panels (b, d, e) report ego-robot costs for \textbf{(S1-3)}. Costs are shown relative to the ground-truth cost. Dashed lines denote the 5th percentile for minimum distances and the 95th percentile for costs. Jointly using image and trajectory observations for Bayesian inverse games (B-PinE-Image-Traj) further improves planning safety in \textbf{(S1)} (larger minimum inter-agent distances and a lower collision rate—0.0\% for B-PinE-Image-Traj vs.\ 1.57\% for B-PinE-Traj) without sacrificing planning efficiency (comparable cost). Moreover, B-PinE-Image-Traj improves motion comfort by avoiding unnecessary steering effort; cf.\ \cref{fig:color-steering}.
  }
  \label{fig:color-quantitative}
\end{figure*}

\begin{figure*}
    \centering
    \includegraphics[width=\linewidth]{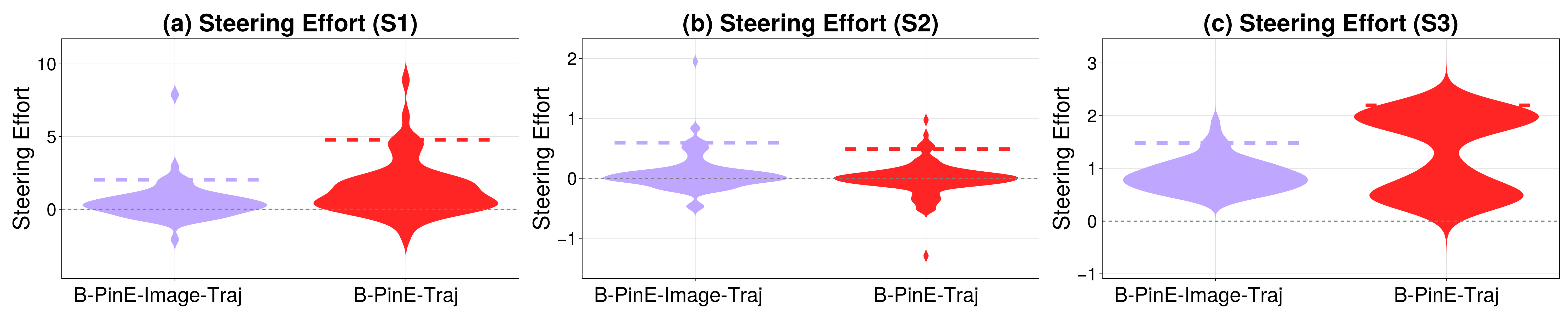}
  \caption{
Steering effort when solving Bayesian inverse games with image-trajectory observations versus trajectory observations alone, shown relative to solving ground truth games. By integrating multi-modal observations, the image-trajectory \ac{vae} disambiguates the opponent’s intent early and avoids unnecessary steering—moderately in \textbf{(S1)} and substantially in \textbf{(S3)}. Dashed lines denote the 95th percentiles. 
  }
  \label{fig:color-steering}
\end{figure*}

\emph{This interaction supports our hypotheses H4-5 and highlights two key advantages of our Bayesian inverse game framework when integrating multimodal observations for inference:}
\begin{itemize}
    \item it can immediately form a posterior by exploiting contextual prior information from the visual modality
    \item it can disambiguate intent in cases where trajectory information alone is insufficient and would otherwise lead to high uncertainty.
\end{itemize}
Together, these properties enable \emph{safer downstream decision-making} in highly dynamic multi-agent interactions that evolve over split seconds.

\paragraph{Monte Carlo Evaluation.}

Next, we quantitatively evaluate the two variants of our Bayesian inverse game framework in a Monte Carlo study. Specifically, we compare the two trained \acp{vae} when coupled with B-PinE as the downstream planner. Each method is run for 500 simulations, with initial conditions and opponent intents randomized in the same manner as in the training data described above.

In total, we draw two main observations from the Monte Carlo study:
\begin{itemize}
    \item \textbf{Planning efficiency and safety.} Planning with our image-trajectory \ac{vae} further improves downstream planning safety relative to the trajectory-only \ac{vae}, without sacrificing planning efficiency as reflected by game costs. Overall, across most runs, planning with either \ac{vae} variant of our Bayesian inverse game approach yields comparable cost performance, and both remain close to the ground truth game cost that assumes known agent intents.
    \item \textbf{Steering effort.} In cases where the opponent drives straight, the image-trajectory \ac{vae} disambiguates the opponent’s intent \emph{early}, reducing unnecessary steering effort and improving motion comfort.
\end{itemize}

\begin{figure*}
  \centering

  \begin{subfigure}{0.374\textwidth}
    \centering
    \includegraphics[width=\linewidth]{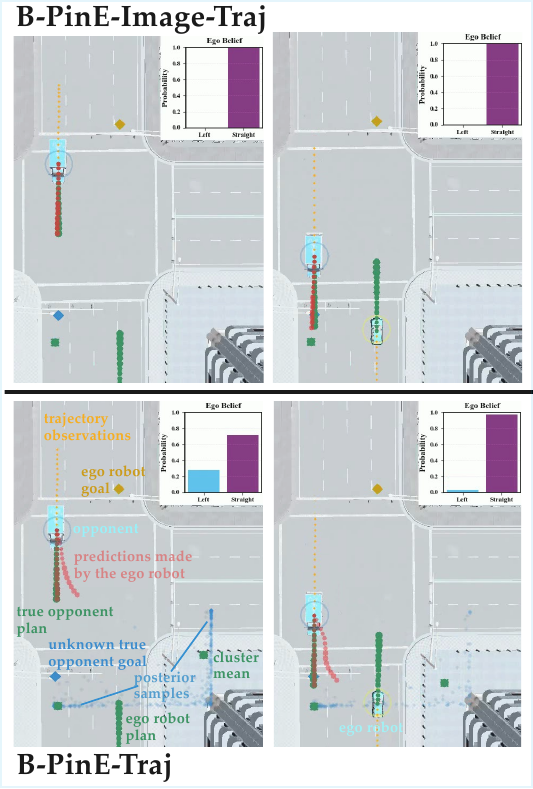}
    \caption{Opponent agent is a truck.}
    \label{fig:truck-qual}
  \end{subfigure}
  \hfill
  \begin{subfigure}{0.55\textwidth}
    \centering
    \includegraphics[width=\linewidth]{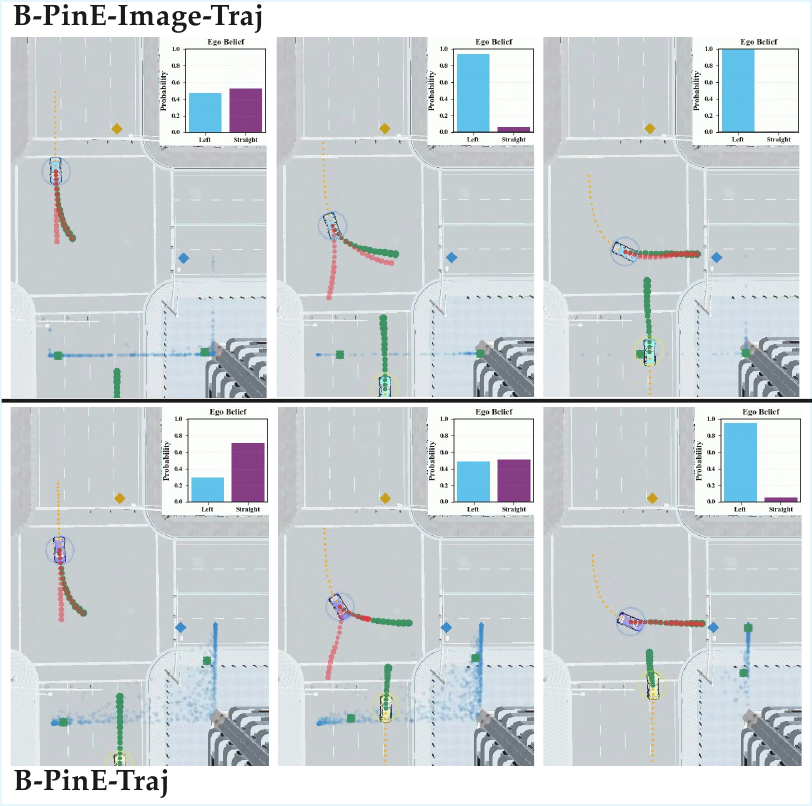}
    \caption{Opponent agent is a car.}
    \label{fig:car-qual}
  \end{subfigure}

  \caption{
Snapshots from an intersection scenario (\cref{sec:type-img}) in which the opponent is either a truck or a car; trucks are forbidden from turning left due to a height limit. When the opponent is a truck, the image-trajectory \ac{vae} immediately infers the opponent’s intent (going straight) by leveraging vehicle-type cues in the visual input, whereas the trajectory-only \ac{vae} recognizes this intent only later from the observed motion (after the opponent passes the intersection). When the opponent is a car, the two \acp{vae} behave similarly, initially producing a bimodal posterior that collapses to a unimodal posterior as the interaction unfolds. 
  }
  \label{fig:truck-example-qualitative}
\end{figure*}

We group the Monte Carlo trials in the same way as in \cref{sec:intersection-state-obs}. \Cref{fig:color-quantitative} reports the cost and minimum inter-agent distance results for \textbf{(S1)} (a–b), \textbf{(S2)} (c–d), and \textbf{(S3)} (e).

Overall, efficiency (cost) and safety (minimum inter-agent distance) are comparable between the image-trajectory and trajectory \acp{vae}. Nonetheless, in \textbf{(S1)} (\cref{fig:color-quantitative}a), the image-trajectory \ac{vae} often achieves a larger minimum inter-agent distance than the trajectory-only \ac{vae} (cf. the higher 5th percentile for the image-trajectory \ac{vae}, denoted by the dashed line), indicating improved safety without sacrificing planning efficiency (as reflected by comparable planning cost). This improved safety margin directly translates to lower collision rates: over the Monte Carlo study, using the same collision-distance threshold as in \cref{sec:intersection-state-obs}, the collision rates for \textbf{(S1)} are 0.0\% for B-PinE-Image-Trajectory and 1.57\% for B-PinE-Trajectory, and in \textbf{(S2)} 0.0\% for both B-PinE-Image-Trajectory and B-PinE-Trajectory. 
We recall that \textbf{(S1)} is more safety-critical where the ego enters the intersection after an aggressive opponent and has to identify the opponent's intent to yield.

In the game costs shown in \cref{fig:color-quantitative}, task-related terms (e.g., distance-to-goal and proximity penalties) strictly dominate the control-effort penalties; cf. \cref{eq:game-cost}. However, for autonomous driving, motion comfort from the passengers’ perspective is equally important, if not more so. 
\Cref{fig:color-steering} shows that B-PinE equipped with the image-trajectory \ac{vae} avoids unnecessary steering maneuvers more effectively than B-PinE with the trajectory-only \ac{vae}—moderately in \textbf{(S1)} (\Cref{fig:color-steering}a) and substantially in \textbf{(S3)} (\Cref{fig:color-steering}c), where the opponent drives straight without interacting with the ego vehicle.
This is intuitive: for the trajectory-only \ac{vae}, before its uncertainty over the opponent’s intent reduces, it tends to act conservatively and hedge against both possibilities (opponent going straight vs.\ turning left). In contrast, the image-trajectory \ac{vae} correctly identifies the opponent’s intent early on and comfortably drives straight  without unnecessary steering.

We emphasize that the conservativeness of B-PinE with the trajectory-only \ac{vae} here does not contradict its aggressiveness in \cref{fig:color-qualitative}. Precisely because the image-trajectory \ac{vae} infers the opponent’s intent early, it can be conservative when the opponent actually turns left and, at the same time, drive straight with ease when the opponent is also going straight.

\emph{Taken together, these results support our hypothesis H6.}

\subsubsection{Agent Type Encoded Intents}
\label{sec:type-img}

\begin{figure*}
    \centering
    \includegraphics[width=\linewidth]{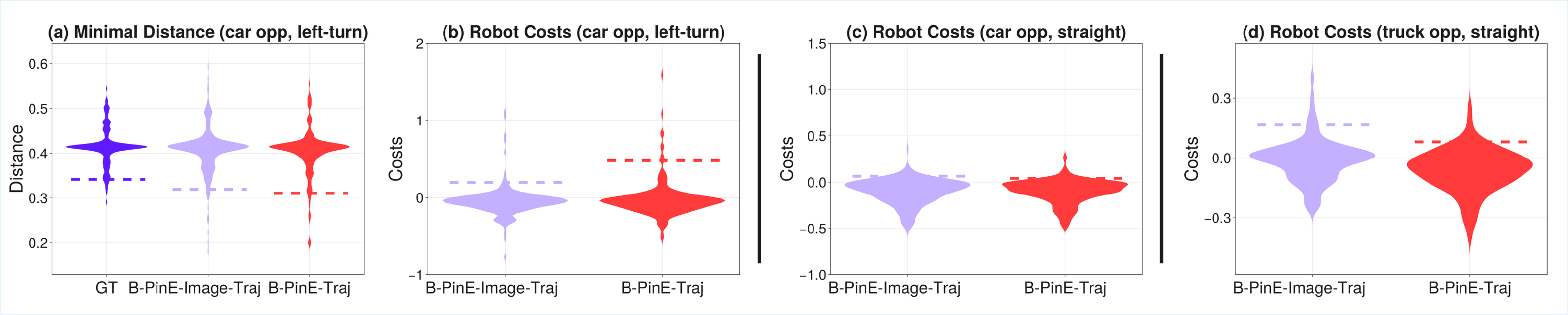}
  \caption{
Monte Carlo results for the intersection scenario in \cref{fig:truck-example-qualitative} (\cref{sec:type-img}). Dashed lines denote the 5th percentile for minimum distances and the 95th percentile for costs. 
Panels (a,b) show the case where the opponent is a left-turning car; panel (c) shows the case where the opponent is a straight-driving car; panel (d) shows the case where the opponent is a straight-driving truck. Costs are shown relative to the ground-truth cost. Across scenarios, the two variants of our approach achieve similar efficiency performance (costs), while incorporating image-trajectory observations improves safety in (a) (collision rates: 0.58\% for B-PinE-Image-Traj vs.\ 1.73\% for B-PinE-Traj). Moreover, B-PinE-Image-Traj improves motion comfort by avoiding unnecessary steering when the opponent is a truck constrained to drive straight; cf.\ \cref{fig:truck-quantitative-b}. 
}
  \label{fig:truck-quantitative-a}
\end{figure*}

\begin{figure}
    \centering
    \includegraphics[width=\linewidth]{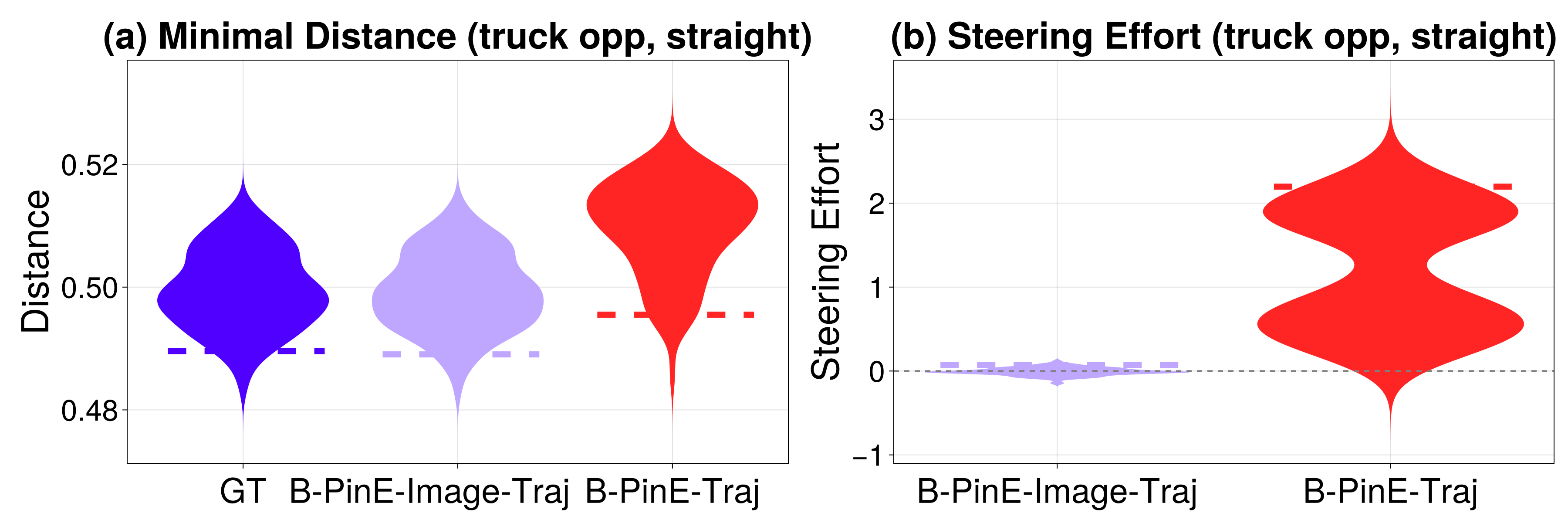}
  \caption{
Minimum distance and steering effort of the two variants of our Bayesian inverse game approach when the opponent is a straight-driving truck (\cref{sec:type-img}). Steering effort is shown relative to solving ground truth games. Dashed lines denote the 5th percentile for minimum distance and the 95th percentile for steering effort. By inferring the opponent’s intent (driving straight) early from image observations of the opponent type, the image-trajectory \ac{vae} improves motion comfort by avoiding unnecessary steering. In contrast, the trajectory-only \ac{vae} remains more cautious—turning away from the oncoming opponent (yielding larger minimum distances) until the truck passes the intersection—and thus incurs higher steering effort.
}
  \label{fig:truck-quantitative-b}
\end{figure}

\Cref{sec:color-img} considers an example where image observations are highly informative about an opponent agent’s intent, corresponding to scenarios such as inferring a driver’s intent from turn signals. Here, we study a setting in which images contain less direct information, but still aid inference of the opponent’s intent.

We evaluate the same intersection scenario as in \cref{sec:color-img}, but randomize the opponent’s color so that it provides no information about intent. Instead, intent can be partially inferred from the opponent’s vehicle type: the opponent is either a car or a truck. In this example, trucks are not allowed to turn left due to a height limit. We construct the training dataset such that the opponent is a car with $80\%$ probability and a truck with $20\%$ probability. When the opponent is a truck, it can only drive straight; when the opponent is a car, it turns left or drives straight with $50\%$ probability each. Aside from these changes, we use the same \ac{vae} training setup as in \cref{sec:color-img}.

In this setting, a well-trained image-trajectory \ac{vae} should immediately infer that a truck opponent will go straight, while producing qualitatively similar inference to the trajectory-only \ac{vae} when the opponent is a car.

\paragraph{Qualitative Behavior.} 
As shown in \cref{fig:truck-example-qualitative}, when the opponent vehicle is a truck (left)—which is forbidden from turning left due to a height limit—the image-trajectory \ac{vae} immediately infers its intent (going straight) by leveraging vehicle type cues in the visual input. This is because, in our interaction dataset, trucks never turn left at this intersection, and this constraint is only observable from the image modality. In contrast, the trajectory-only \ac{vae} produces a bimodal posterior initially, which collapses to (nearly) unimodal only after the opponent has almost passed the intersection. 

When the opponent vehicle is a car (\cref{fig:truck-example-qualitative}, right), the two \acp{vae} behave similarly: both infer a bimodal posterior early on that collapses to a unimodal posterior as the interaction unfolds, reflecting increasing certainty about the opponent’s intent.

The trajectory-only \ac{vae} produces an imbalanced initial posterior because it cannot observe vehicle type; since trucks can only go straight, the overall probability of going straight is higher in this setting.

\textit{These results support our hypotheses H4-5 and highlight a useful feature of our Bayesian inverse game framework}: \emph{it can integrate prior information across observation modalities}. When images provide additional informative cues, the framework leverages them to complement trajectory-based inference; when they provide little extra information, it naturally falls back to relying primarily on trajectory observations.

\paragraph{Monte Carlo Evaluation.}

Next, we quantitatively evaluate the two variants of our Bayesian inverse game framework via a Monte Carlo study. As in \cref{sec:color-img}, we compare the two trained \acp{vae} when coupled with B-PinE as the downstream planner. We run each method for 500 simulations, randomizing the initial conditions, opponent types, and opponent intents as described above.

We group the trials into three scenarios: (i) the opponent is a left-turning car, (ii) the opponent is a straight-driving car, and (iii) the opponent is a straight-driving truck. In (i) and (ii), we expect the two \ac{vae} variants to perform similarly, since observing that the opponent is a car does not disambiguate its intent for the image-trajectory \ac{vae}. In contrast, in (iii), the image-trajectory \ac{vae} can infer the opponent’s intent early—because trucks are constrained to drive straight due to a height limit—and should therefore behave less conservatively (e.g., requiring less steering).

\Cref{fig:truck-quantitative-a} reports the minimum distances and ego-robot costs. Overall, the two \ac{vae} variants achieve comparable efficiency as measured by robot cost, while the image-trajectory \ac{vae} improves safety relative to the trajectory-only \ac{vae}: in trials where the opponent car turns left, the collision rates are 0.58\% for B-PinE-Image-Traj versus 1.73\% for B-PinE-Traj.

Moreover, \cref{fig:truck-quantitative-b} zooms in on the case where the opponent is a straight-driving truck and reports the minimum inter-agent distance and steering effort for the two variants. Consistent with the qualitative results in \cref{fig:truck-example-qualitative}, the image-trajectory \ac{vae} infers the opponent’s intent early and behaves less conservatively, resulting in a smaller minimum distance and lower steering effort. In contrast, the trajectory \ac{vae} typically identifies the intent only after the truck passes the intersection (cf.\ \cref{fig:truck-example-qualitative}); consequently, B-PinE-Traj turns away from the oncoming opponent despite the lack of interaction, leading to larger minimum distances and higher steering effort. Overall, early intent identification via multi-modal observations improves motion comfort in this setting.

The quantitative results are similar to those in \cref{sec:color-img} and \textit{support our hypothesis H6}.

\section{Conclusions}\label{sec:conclusions}

We studied \emph{inverse games} for interactive decision-making under incomplete game models. We introduced a structured \ac{vae} framework that approximately solves Bayesian inverse games and infers full posterior distributions over unknown game parameters (e.g., opponent objectives) from multimodal observations, including high-dimensional images. The framework embeds a differentiable Nash game solver within the \ac{vae}, providing an inductive bias that preserves interpretability and grounds posterior samples in the game parameter space.

We extensively evaluated the proposed approach in interactive decision-making scenarios. Compared with common \ac{mle}-based inverse game methods that yield only point estimates of game parameters, our framework provides meaningful uncertainty quantification, effectively leverages prior information available in offline datasets, and enables safer downstream decision-making in interactive motion planning. After training the structured \ac{vae} offline on interaction data without objective labels, the model supports real-time posterior sampling at test time without additional game solves, making it practical for online decision-making. Finally, by learning a shared latent space across modalities, the proposed approach can exploit contextual visual information when trajectory history is limited—e.g., when a new agent enters the scene and little or no past behavior is available.

Several directions could further extend this work. First, our current implementation uses a decoder that maps the latent variable $z$ deterministically to game parameters $\theta$, consistent with the standard \ac{vae} formulation~\citep{kingma2013auto}. A natural extension is to introduce a stochastic decoding mechanism—e.g., via hierarchical latent-variable models or diffusion-based generative models—to more fully leverage the expressive power of modern generative architectures. Second, the modular structure of our pipeline enables studying alternative equilibrium concepts, such as entropic cost equilibria~\citep{mehr2023maxent}, which can capture bounded-rationality assumptions.

\section*{Author Contributions}
\begin{description}
    \item[Yash Jain:] Methodology, Data curation, Investigation, Software, Writing -- Original Draft, Writing -- review \& editing
    \item[Xinjie Liu:] Conceptualization, Data curation, Formal analysis,  Investigation,  Methodology, Software, Writing -- Original Draft, Writing -- review \& editing
    \item[Lasse Peters:] Conceptualization, Formal analysis, Methodology, Writing -- Original Draft, Writing -- review \& editing
    \item[David Fridovich-Keil:] Funding acquisition, Conceptualization, Supervision, Writing -- Original Draft, Writing -- review \& editing
    \item[Ufuk Topcu:] Funding acquisition, Conceptualization, Supervision, Writing -- Original Draft, Writing -- review \& editing
\end{description}

\section*{Statements and Declarations}
    
\subsection*{Ethical considerations}

This article does not contain any studies with human or animal participants; all experiments were conducted in simulation. 

\subsection*{Consent to participate}

Not applicable.

\subsection*{Consent for publication}

Not applicable.

\subsection*{Declaration of conflicting interest}

The authors declared no potential conflicts of interest with respect to the research, authorship, and/or publication of this article.

\subsection*{Funding statement}

This research was sponsored by the Army Research Laboratory under Cooperative Agreements W911NF-23-2-0011, W911NF-25-2-0021, and W911NF-23-S-0001, and by the National Science Foundation under Grants 2211432, 2211548, and 2336840.

\subsection*{Data availability}

The implementation of the differentiable game solver is publicly available~\citepalias{MCPTrajectoryGameSolver}, and the experiment code needed to reproduce the results will be released under the same GitHub organization.

\subsection*{Use of AI tools}
During the preparation of this manuscript, the authors used ChatGPT (OpenAI) and Gemini (Google) as assistive tools to improve language clarity, grammar, and overall writing quality, and to help review and debug research software (e.g., identifying potential coding errors and suggesting improvements). The authors critically reviewed and edited all AI-assisted outputs, verified technical correctness and any cited material as needed, and take full responsibility for the accuracy, validity, and originality of the manuscript and its references. The authors further acknowledge the limitations of large language models, including the potential for bias, errors, or omissions, and confirm that no AI-generated content was included without appropriate human verification.

\bibliographystyle{SageH}
\bibliography{main.bib}

\end{document}